\documentclass[lettersize,journal]{IEEEtran}
\usepackage{amsmath,amsfonts}
\usepackage{algorithm}
\usepackage{array}
\usepackage[caption=false,font=normalsize,labelfont=sf,textfont=sf]{subfig}
\usepackage{textcomp}
\usepackage{stfloats}
\usepackage{url}
\usepackage{verbatim}
\usepackage{graphicx}
\usepackage{adjustbox}
\usepackage{cite}
\usepackage{amsmath}
\usepackage{amssymb}
\usepackage{booktabs}
\usepackage{times}
\usepackage{soul}
\usepackage{inputenc}
\usepackage{amsmath}
\usepackage{amsthm}
\usepackage{amsfonts}
\usepackage{booktabs}
\usepackage[switch]{lineno}
\usepackage{color}
\usepackage{multirow}
\usepackage{float}
\usepackage{colortbl}
\usepackage[colorlinks,linkcolor=red,anchorcolor=black,citecolor=green]{hyperref}
\usepackage{algorithmicx}
\usepackage{algpseudocode}
\usepackage[table,xcdraw]{xcolor}

\usepackage{caption}
\usepackage{subcaption}
\usepackage{tikz}
\usepackage{caption}
\usepackage{xcolor} 
\definecolor{clear sky}{RGB}{79,253,199}
\definecolor{thick cloud}{RGB}{77,2,115}
\definecolor{thin cloud}{RGB}{251,255,41}
\definecolor{cloud shadow}{RGB}{221,53,223}

\begin{document}

\title{Adapting Vision Foundation Models for Robust Cloud Segmentation in Remote Sensing Images}

\author{
Xuechao Zou,~\IEEEmembership{Student Member, IEEE}, Shun Zhang, Kai Li,~\IEEEmembership{Student Member, IEEE}, Shiying Wang, Junliang Xing,~\IEEEmembership{Senior Member, IEEE}, Lei Jin, Congyan Lang,~\IEEEmembership{Member, IEEE}, and Pin Tao,~\IEEEmembership{Member, IEEE}

\thanks{ 
This work was partly supported by the National Natural Science Foundation of China under Grant 62072027. Xuechao Zou, Shun Zhang and Kai Li have contributed equally to this work. Corresponding authors: Congyan Lang and Pin Tao.

Xuechao Zou and Congyan Lang are with the Key Lab of Big Data \& Artificial Intelligence in Transportation (Ministry of Education), School of Computer Science \& Technology, Beijing Jiaotong University, Beijing, China (e-mail: xuechaozou@foxmail.com; cylang@bjtu.edu.cn).

Kai Li, Junliang Xing, and Pin Tao are with the Department of Computer Science and Technology, Tsinghua University, Beijing, China. Pin Tao is also with the Key Laboratory of Pervasive Computing, Ministry of Education (e-mail: tsinghua.kaili@gmail.com; jlxing@tsinghua.edu.cn; taopin@tsinghua.edu.cn).

Shun Zhang and Shiying Wang are with the School of Computer Technology and Application, Qinghai University, Xining, China (e-mail: xiaoshun3238@gmail.com; wangshiying.qhu@foxmail.com).

Lei Jin is with the School of Electronic Engineering, Beijing University of Posts and Telecommunications, Beijing, China (e-mail: jinlei@bupt.edu.cn).}
}


\maketitle

\begin{abstract}
Cloud segmentation is a critical challenge in remote sensing image interpretation, as its accuracy directly impacts the effectiveness of subsequent data processing and analysis. Recently, vision foundation models (VFM) have demonstrated powerful generalization capabilities across various visual tasks. In this paper, we present a parameter-efficient adaptive approach, termed Cloud-Adapter, designed to enhance the accuracy and robustness of cloud segmentation. Our method leverages a VFM pretrained on general domain data, which remains frozen, eliminating the need for additional training. Cloud-Adapter incorporates a lightweight spatial perception module that initially utilizes a convolutional neural network (ConvNet) to extract dense spatial representations. These multi-scale features are then aggregated and serve as contextual inputs to an adapting module, which modulates the frozen transformer layers within the VFM. Experimental results demonstrate that the Cloud-Adapter approach, utilizing only 0.6\% of the trainable parameters of the frozen backbone, achieves substantial performance gains. Cloud-Adapter consistently achieves state-of-the-art performance across various cloud segmentation datasets from multiple satellite sources, sensor series, data processing levels, land cover scenarios, and annotation granularities. We have released the code and model checkpoints at \url{https://xavierjiezou.github.io/Cloud-Adapter/} to support further research.
\end{abstract}


\section{Introduction}\label{intro}

\begin{figure}[!t]
\centering
\includegraphics[width=0.95\linewidth]{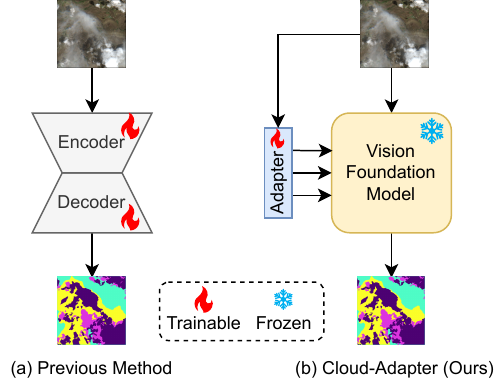}
\caption{Comparison between (a) the Previous Method and (b) our Cloud-Adapter approach. Unlike (a), where the entire network (e.g., a typical encoder-decoder architecture) is fully trainable, resulting in a large number of parameters and an increased risk of overfitting, (b) our Cloud-Adapter approach leverages a frozen vision foundation model (VFM) combined with a lightweight, trainable adapter. This design preserves generalization and adaptability, enabling efficient learning for cloud segmentation tasks. The frozen VFM extracts generic visual features, while the parameter-efficient adapter facilitates effective transfer learning on diverse remote sensing scenes.}
\label{fig: pipeline}
\end{figure}

\IEEEPARstart{C}{loud} segmentation, also known as cloud detection, plays a critical role in the interpretation of remote sensing images, with wide applications in climate monitoring~\cite{climate_monitoring_1, climate_monitoring_2}, environmental protection, disaster warning, and land use~\cite{land_use}. The variations and characteristics of clouds significantly influence the quality of remote sensing images and subsequent analysis results. Therefore, accurate cloud segmentation is essential for extracting meaningful geospatial information and ensuring reliable data processing. However, due to the diversity of cloud shapes, textures, and sizes, coupled with the complexity of imaging conditions, traditional cloud segmentation methods~\cite{hot, ihot, fmask, cfmask} often fail to achieve stable high-accuracy results in dynamic remote sensing images. Consequently, developing a more robust and adaptable cloud segmentation method has become a key research focus in remote sensing.

With the rapid development of deep learning~\cite{deeplearning, lenet, alexnet, googlenet}, convolutional neural network (ConvNet)-based cloud segmentation methods~\cite{hrc_whu, rsnet, mfcnn, 7729176, cdnetv1, cdnetv2, dabnet} have made remarkable progress, particularly due to their ability to automatically extract high-level features that enhance the accuracy of cloud shape, edge detection, and texture analysis. For instance, MFCNN~\cite{mfcnn} combines multi-scale convolutional features to enable fine-grained cloud detection. At the same time, DABNet~\cite{dabnet} introduces the deformable context feature pyramid module to improve the adaptive modeling capability of multi-scale features for high-accuracy cloud detection. Despite these advancements, many methods still face challenges, including strong dependence on training data, low model generalization, and limited adaptability to diverse remote sensing scenes.

Meanwhile, vision foundation models (VFM)~\cite{vit, swin, mae, sam, dinov2}, which learn task-agnostic representations from large-scale datasets~\cite{imagenet}, have demonstrated impressive performance across various applications, such as image classification~\cite{crossvit}, object detection~\cite{sam-cod}, and segmentation~\cite{st-unet, maskdino, semmae}. Transformer architectures like ViT~\cite{vit} and Swin Transformer~\cite{swin} have become dominant in the VFM landscape, showing superior performance. However, the potential of VFMs for cloud segmentation remains largely unexplored.

Existing cloud segmentation methods~\cite{mcdnet, dbnet} primarily rely on complex neural network architectures, most of which require extensive retraining or fine-tuning on large labeled datasets. Although these methods achieve good performance, they suffer from several limitations. First, they are computationally expensive and require a large amount of labeled data, which is often impractical for real-world applications. Second, traditional models are prone to overfitting, particularly when dealing with small datasets or high-resolution remote sensing images. Third, most existing methods~\cite{mcdnet, scnn, kappamask, cdnetv1, cdnetv2, dbnet, hrcloudnet} have barely leveraged the power of VFM for cloud segmentation, missing the opportunity to exploit their robust transfer learning capabilities with minimal additional parameters.

To address these challenges, we propose a novel cloud segmentation method based on vision foundation models named Cloud-Adapter (see Fig.~\ref{fig: pipeline}). This approach freezes the pretrained VFM backbone and incorporates a lightweight adaptive module, requiring minimal additional parameters to enhance cloud segmentation accuracy and robustness. Specifically, we design a spatial perception module that employs convolutional neural networks (ConvNets) to extract dense spatial features and aggregate them into multi-scale contextual information. This information is then passed into the adaptive module, which modulates the frozen transformer layers of the VFM to perform cloud segmentation efficiently. Experimental results demonstrate that Cloud-Adapter achieves significant performance improvements with only 0.6\% of the trainable parameters of the backbone across various popular datasets.

The contributions of this work are summarized as follows:

\begin{enumerate}
    \item We propose Cloud-Adapter, a simple but powerful parameter-efficient fine-tuning method for cloud segmentation that freezes the backbone of VFMs and fine-tunes only a small fraction of the frozen parameters.
    \item We introduce two lightweight components: the spatial perception module and the adapting module, which modulate the features of the VFM, enabling adaptation from a general domain to a remote sensing domain.
    \item We demonstrate that our method consistently achieves SOTA performance on four datasets (six subsets), including multiple satellite sources, sensor types, and land cover scenarios, highlighting its robustness towards remote sensing cloud segmentation tasks.
\end{enumerate}

\begin{figure*}[!t]
\centering
\includegraphics[width=0.95\linewidth]{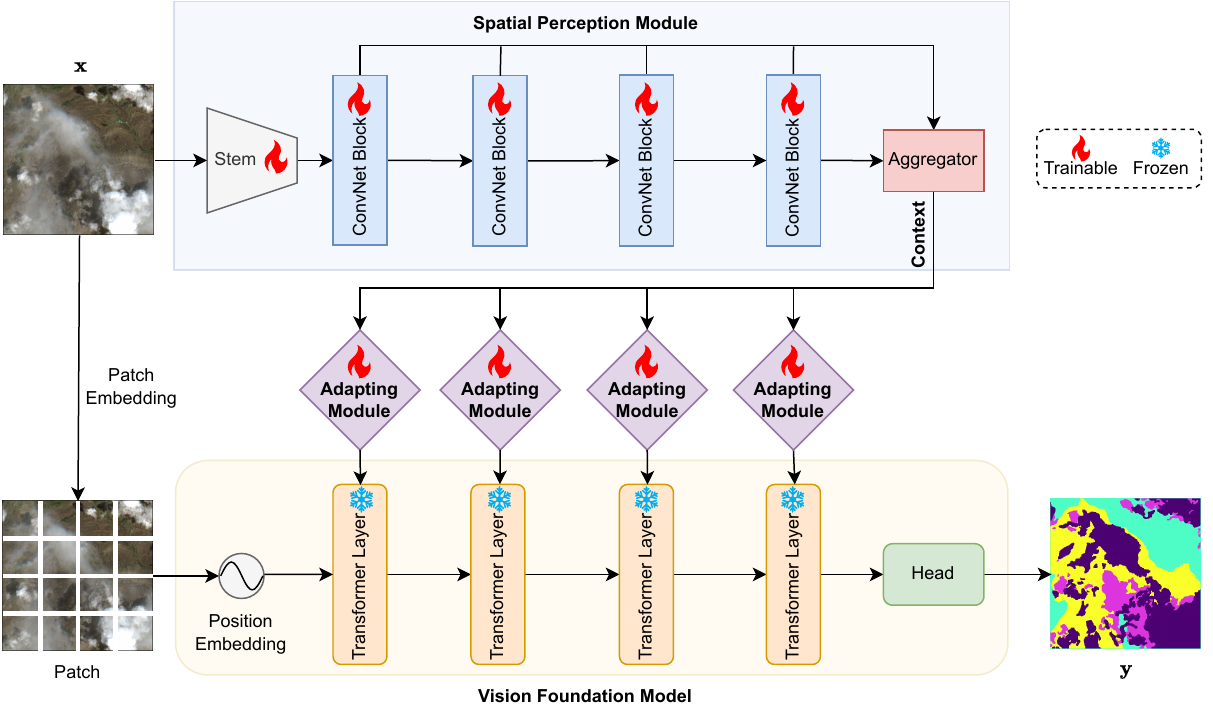}
\caption{Detailed network architecture of the proposed Cloud-Adapter method, consisting of the spatial perception and adapting modules. The spatial perception module uses ConvNet blocks to extract dense spatial features, which are aggregated into a multi-scale context and fed to the adapting module. The adapting module modulates the frozen transformer layers in the VFM.}
\label{fig: framework}
\end{figure*}

\section{Related Work}\label{sec: relatedwork}

\subsection{Cloud Segmentation}
\noindent{Early cloud segmentation methods primarily relied on the spectral and statistical features of images, using techniques like thresholding, multispectral fusion, and edge detection to distinguish clouds from other surface elements. For instance, CFMask~\cite{fmask, cfmask} employs decision trees to label pixels and projects estimated cloud heights onto the ground to generate a cloud shadow mask. HOT~\cite{hot, ihot} introduces a transformation optimized for detecting cloud and haze distributions in Landsat images, amplifying haze variations via spectral sensitivity.}

However, these traditional methods are highly sensitive to cloud shapes, lighting conditions, and atmospheric environments, resulting in suboptimal performance in complex remote sensing scenarios. With advancements in deep learning, ConvNet-based cloud segmentation methods have gained significant attention due to their ability to automatically extract high-level features, enhancing the accuracy of cloud shape, texture, and edge detection. Examples of innovative model architectures include MSCFF~\cite{hrc_whu}, which combines multi-scale convolutional features to facilitate cloud detection. DBNet~\cite{dbnet} employs a hybrid model integrating Transformers and ConvNets to capture semantic and spatial details, thereby reducing false detections and omissions in cloud segmentation. HRCloudNet~\cite{hrcloudnet} adopts a hierarchical integration approach to preserve cloud texture in high-resolution images, while MCDNet~\cite{mcdnet} achieves effective thin cloud segmentation through multi-scale feature fusion and auxiliary cloud removal.

For models designed for specific scenarios and training needs, CDNetv1~\cite{cdnetv1} leverages a feature pyramid and a boundary refinement block to extract cloud masks from low-resolution thumbnails. CDNetv2~\cite{cdnetv2} improves performance in mixed cloud-snow scenes using adaptive feature fusion and high-level semantic guidance. SCNN~\cite{scnn} applies a shallow ConvNet with filters of varying sizes to minimize model training costs. Additionally, TransMCD~\cite{gf12ms_whu} uses pseudo-labels derived from spectral features, enabling weakly supervised cloud detection and enhancing model generalization.

Despite these advancements, achieving robust segmentation performance across diverse remote sensing images remains a significant challenge. Furthermore, most methods continue to rely on traditional network architectures, such as UNet~\cite{unet}, ResNet~\cite{resnet}, FPN~\cite{fpn}, and DeepLab~\cite{deeplab}, with limited exploration of vision foundation models for cloud segmentation.

\subsection{Vision Foundation Models}
\noindent{Learning task-agnostic pre-trained representations~\cite{bert, gpt3} has become a standard practice in natural language processing. Inspired by this paradigm, substantial progress has been achieved in exploring vision foundation models (VFMs). Typically, these models are trained on large-scale datasets to learn general-purpose representations, achieving remarkable performance across diverse applications, including image classification, object detection, and segmentation. Transformer architectures~\cite{vit, swin} dominate the landscape of vision foundation models, demonstrating superior performance. CLIP~\cite{clip} aligns text and image modalities in latent space through contrastive learning. MAE~\cite{mae} learns robust representations via self-supervised masked modeling. SAM~\cite{sam} incorporates interactive prompts to capture general concepts of objects, while DINOv2~\cite{dinov2} employs self-distillation to learn generic semantic features. These models effectively capture complex visual patterns and exhibit strong capabilities for few-shot and even zero-shot transfer, allowing them to adapt well to new tasks. Given the diversity and complexity of remote sensing images, exploring vision foundation models with strong generalization is essential for achieving robust cloud segmentation.}

\subsection{Transfer Learning and Domain Adaptation}
\noindent{In resource-constrained environments or when dealing with large-scale data, training models from scratch is often impractical and costly. Transfer learning~\cite{transfer-learning} enables adapting models trained on large datasets to specific tasks. This approach typically involves fine-tuning pre-trained models on smaller, domain-specific datasets, allowing the model to retain general knowledge while learning domain-specific characteristics. Domain adaptation~\cite{domain-adaptation} further enhances this process by aligning the feature distributions between the source domain (where the model was initially trained) and the target domain (the new application environment), thus maximizing task performance.}

Adapters exemplify the integration of transfer learning and domain adaptation, offering a generally plug-and-play approach across various tasks. Initially introduced in natural language processing (NLP)~\cite{adapter}, chapters have rapidly evolved. In NLP tasks, LoRA~\cite{lora} injects trainable low-rank decomposition matrices into each layer of the Transformer architecture, enabling large language models to adapt to specific domains. In multimodal generation tasks, IP-Adapter~\cite{ip-adapter} is an effective adapter for implementing image prompting functions in pre-trained text-to-image diffusion models. T2I-Adapter~\cite{t2i-adapter} introduces a lightweight adapter to modulate the intensity of internal knowledge and external control signals in text-to-image models. For visual segmentation tasks, ViT-Adapter~\cite{vit-adapter} integrates image-related inductive biases into pre-trained ViT models, improving dense prediction performance. SAM-Adapter~\cite{sam-adapter} introduces a simple adapter that incorporates domain-specific information or visual prompts into SAM~\cite{sam}, enhancing performance in specific scenarios. Therefore, developing an efficient adapter shows great potential for improving the robustness of cloud segmentation by leveraging the strengths of vision foundation models.

\section{Method}\label{sec: cloud-adapter}

\noindent{In this section, we introduce a simple but powerful adaptive method called Cloud-Adapter (as shown in Fig.~\ref{fig: framework}) to achieve fine-tuning of cloud segmentation by freezing the weights of the visual foundation model (VFM). The method consists of two key components: the spatial perception module and the adapting modules. The spatial perception module extracts multi-scale spatial features using multiple ConvNet blocks and then aggregates them into a dense context. The adapting modules modulate the frozen VFM layers with the context, enabling effective transfer learning for cloud segmentation.}

\subsection{Spatial Perception Module}

The spatial perception module (SPM) extracts multi-scale spatial features from the input remote sensing images. First, the input image goes through a stem module for preliminary feature extraction. The stem consists of two depthwise separable convolutions~\cite{dwsc}. The first convolution uses a $1 \times 1$ kernel to map the input image’s channels to a larger feature space, primarily for channel-wise information fusion. The second convolution uses a $3 \times 3$ kernel to extract spatial features and capture local spatial information.

Let the input image be $\mathbf{x} \in \mathbb{R}^{C \times H \times W}$, where $C$ is the number of channels, and $H$ and $W$ are the height and width of the image. The output feature map after stem is $\mathbf{F}_{\text{stem}} \in \mathbb{R}^{C' \times H \times W}$, where $C'$ is the mapped number of channels, and $H$ and $W$ are the feature map dimensions, which remain the same as the input image.

After the stem, the input features are processed by multiple ConvNet blocks, each of which extracts spatial features at different scales, resulting in a series of feature maps $\mathbf{F}_i^{c} \in \mathbb{R}^{C_i \times H_i \times W_i}$, where $i$ denotes different scales. To aggregate these multi-scale feature maps into a unified scale, we use an aggregator, which applies adaptive average pooling to normalize the feature maps to the same size. The aggregator is a parameter-free operation, meaning it does not contain any trainable parameters, making it a zero-parameter operation.

The aggregation of multi-scale features is expressed as:

\begin{equation}
\mathbf{F}_{\text{agg}} = \sum_{i=1}^{k} \mathcal{P}_{avg}(\mathbf{F}_i^{c}),
\end{equation}
where $\mathbf{F}_i^{c}$ are the feature maps at different scales, and $\mathcal{P}_{avg}$ is the adaptive average pooling operation that pools the feature maps to the smallest size of the final feature map $\mathbf{F}_k^{c}$. Additionally, $k \in \{1, 2, \dots, \lfloor \log_2(\min(H_0, W_0)) \rfloor \}$ refers to the number of ConvNet blocks, where each block has a convolution with the stride of 2, achieving a downsampling effect by squeezing the spatial dimensions at each step. Using the aggregator, we integrate multi-scale hierarchical representations into a single dense prior $\mathbf{F}_{\text{agg}} \in \mathbb{R}^{C_k \times H_k \times W_k}$.

\subsection{Adapting Module}

\begin{figure}[!t]
\centering
\includegraphics[width=0.95\linewidth]{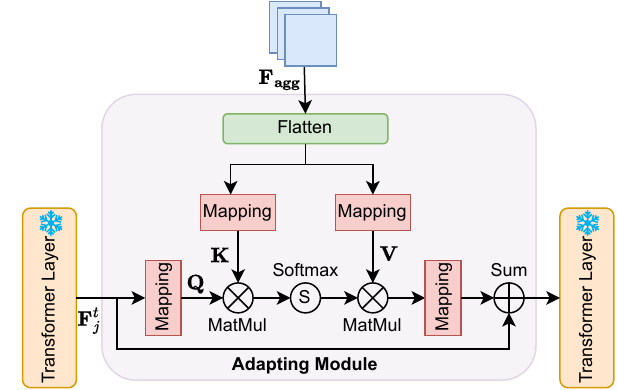}
\caption{Schematic diagram of the proposed adapting module.}
\label{fig: adapting}
\end{figure}

The adapting module, as referred to in Fig.~\ref{fig: adapting}, utilizes the cross-attention mechanisms~\cite{cat, ccnet, u-transformer, pmaa} to adjust the frozen VFM's Transformer layers. Specifically, it takes the context $\mathbf{F}_{\text{agg}}$ from the spatial perception module and interacts with the frozen VFM features, thereby modulating the output of the VFM. The cross-attention mechanism calculates the relationships between aggregated dense features and the features extracted by the VFM, producing the adapted features. The queries, keys, and values are computed as follows:
\begin{equation}
\mathbf{Q} = \mathcal{M}_q(\mathbf{F}_j^t), \quad \mathbf{K} = \mathcal{M}_k(\mathbf{F}_{\text{agg}}), \quad \mathbf{V} = \mathcal{M}_v(\mathbf{F}_{\text{agg}}),
\end{equation}
where $\mathbf{Q} \in \mathbb{R}^{d}$, $\mathbf{K} \in \mathbb{R}^{d}$, and $\mathbf{V} \in \mathbb{R}^{d}$ are the query, key, and value matrices, respectively. \( \mathcal{M}_q \), \( \mathcal{M}_k \), and \( \mathcal{M}_v \) denote learnable linear projection functions for generating the respective matrices. $\mathbf{F}_j^{t}$ represents the robust feature extracted from the frozen VFM transformer layers, where $j$ indicates the index. Then cross-attention computation is expressed by:

\begin{equation}
\mathbf{S} = \text{Softmax}\left( \frac{\mathbf{QK}^T}{\sqrt{d}} \right),
\end{equation}

\begin{equation}
\mathbf{O} = \mathbf{S} \cdot \mathbf{V},
\end{equation}

\begin{equation}
\mathbf{F}_j^{att} = \mathcal{M}(\mathbf{O}) + \mathbf{F}_j^{t},
\end{equation}
where the Softmax operation computes the attention weights $\mathbf{S}$, which are then multiplied with the value matrix $\mathbf{V}$ to obtain the output $\mathbf{O}$. Finally, after the mapping operation (denoted as $\mathcal{M}$), the adapted features $\mathbf{F}_j^{att}$ are obtained with a residual.

The mapping operation is performed via a multi-layer perceptron (MLP) with the low-rank~\cite{lora} design without any activation, aimed to achieve feature projection with minimal parameter cost. The mapping process can be expressed as:

\begin{equation}
\mathcal{M}(\mathbf{z}) = \mathbf{W}_2 \cdot (\mathbf{W}_1 \cdot \mathbf{z}),
\end{equation}
where $\mathbf{W}_1 \in \mathbb{R}^{d \times r}$ and $\mathbf{W}_2 \in \mathbb{R}^{r \times d}$ are the weight matrices of the fully connected layers, $\mathbf{z}$ is the input feature, and $r$ is the rank for dimensionality reduction. The smaller the value of $r$, the fewer trainable parameters the adapting module contains.

\section{Experiments}

\subsection{Datasets}

\noindent{We conducted experiments on four popular cloud segmentation datasets, comprising six subsets, from different satellite sources: CloudSEN12\_High~\cite{cloudsen12_high}, L8\_Biome~\cite{l8_biome}, GF12MS\_WHU~\cite{gf12ms_whu}, and HRC\_WHU~\cite{hrc_whu}, as summarized in Table~\ref{tab: datasets}. Specifically, CloudSEN12\_High contains two data processing levels, L1C and L2A, while GF12MS\_WHU consists of data captured by two satellite sensors, GF1 and GF2. Notably, in typical binary cloud segmentation datasets, thin and thick clouds are grouped into a single category, ``cloud," while cloud shadows are excluded from the annotations.}

\subsubsection{CloudSEN12\_High}~\cite{cloudsen12_high}
CloudSEN12 is a global dataset designed for semantic understanding of clouds and their shadows in Sentinel-2 imagery. It addresses the long-standing challenge of accurately characterizing clouds and cloud shadows in the earth observation community. This dataset comprises 49,400 image patches, including Sentinel-2 Level-1C and Level-2A multispectral data, Sentinel-1 synthetic aperture radar data, auxiliary remote sensing products, and meticulously crafted annotations identifying thick clouds, thin clouds, and cloud shadows. CloudSEN12 stands out due to its diverse annotations, scene variability, broad geographic coverage, and metadata complexity, making it an essential resource for benchmarking and improving cloud detection methodologies. Here, we select its high-quality version with four classes of fine-grained annotations for our experiments.

\subsubsection{L8\_Biome}~\cite{l8_biome}
The L8\_Biome dataset is designed to facilitate comparing and validating cloud detection algorithms for operational Landsat data products. This dataset encompasses various scenario types, including Urban, Barren, Forest, Shrubland, Grass/Cropland, Snow/Ice, Wetlands, and Water. Such a variety of scene types enables models to learn more diverse features, handle complex surface backgrounds, and enhance the accuracy of cloud removal tasks across different terrains. The dataset includes manually generated cloud masks to support the validation of cloud cover assessment algorithms, enabling precise computation of cloud cover percentages for each scene. These characteristics make the dataset an essential resource for cloud segmentation. Here, we cropped each tile into 512 $\times$ 512 patches and removed images that did not contain any targets of interest. The remaining data were then split into training, validation, and test sets in a ratio of 6:2:2.

\subsubsection{GF12MS\_WHU}~\cite{gf12ms_whu}
The GFMS\_WHU dataset consists of 8-meter Gaofen1-MS and 4-meter Gaofen2-MS images. Gaofen1 and Gaofen2 belong to the same satellite series but are equipped with different sensors, offering varying spatial resolutions to enhance the model's generalization capability. These images were collected from diverse regions of China between June 2014 and December 2020, improving the dataset's ability to support robust model training. To standardize input dimensions, all images were uniformly cropped and padded to a size of 256 $\times$ 256 patches. The dataset is split into a training set and a test set with a ratio of 7:3, providing a balanced framework for model development and evaluation.

\subsubsection{HRC\_WHU}~\cite{hrc_whu}
HRC\_WHU is a high-resolution cloud cover validation dataset. It comprises 150 high-resolution images acquired from Google Earth with RGB channels and resolutions varying from 0.5 to 15 meters. These images have been expertly digitized with two mask annotations: clear sky and clouds. Here, We uniform cut to 256 $\times$ 256 patch size, 120 for training, and 30 for testing.

\begin{table*}[ht]
\centering
\caption{Summary of information for four popular cloud segmentation datasets. ``-" indicates non-existence or not mentioned.}
\label{tab: datasets}
\setlength{\tabcolsep}{2.8mm}
\begin{tabular}{l|c|c|ccc|c|c} 
\toprule
\textbf{Dataset Name}                       & \textbf{Satellite Source}                   & \textbf{Ground Resolution} & \textbf{Training Set} & \textbf{Validation Set} & \textbf{Testing Set} & \textbf{Patch Size} & \textbf{Classes} \\
\midrule
CloudSEN12\_High~\cite{cloudsen12_high}                          & Sentinel 2                      & 10m                 & 8,490              & 535                 & 975              & 512$\times$512 & 4           \\ 
L8\_Biome~\cite{l8_biome}                           & Landsat 8                & 30m                 & 7,931                 & 2,643                   & 2,643                & 512$\times$512 & 4  \\ 
GF12MS\_WHU~\cite{gf12ms_whu}                           & Gaofen 1, 2                        & 8m \& 4m                  & 20,697              & 11,645                   & -             & 256$\times$256 & 2            \\ 
HRC\_WHU~\cite{hrc_whu}                            & Google Earth                      & 0.5m-15m          & 120               & -                   & 30               & 256$\times$256 & 2         \\ 
\bottomrule
\end{tabular}
\end{table*}

\subsection{Implementation Details}

\subsubsection{Training Settings}
All experiments were conducted on a workstation with NVIDIA GeForce RTX 3090 GPUs. The number of ConvNet blocks ($k$) was set to 4. We employed a custom learning rate schedule to optimize model performance. The schedule began with a warm-up phase, utilizing a linear scheduler that increased the learning rate from $1 \times 10^{-6}$ over the first 1,000 iterations. This was followed by a PolyLR scheduler with an eta minimum of 0.0 and a power of 0.9, applied from iteration 1,000 to 40,000. The training loop was configured to run for a maximum of 40,000 iterations, with validation conducted at intervals of every 4,000 steps. To ensure reproducibility, we set the random seed to 42 and used the AdamW optimizer~\cite{adamw} with an initial learning rate of $1 \times 10^{-4}$ and a weight decay of 0.05. The batch size during training was set to 4, and parameters were initialized using the kaiming normal~\cite{kaiming}. The segmentation head used in this study was the widely-used Mask2Former~\cite{mask2former} to generate masks.

\subsubsection{Evaluation Metrics}
To comprehensively evaluate cloud segmentation performance, we adapt several widely recognized metrics: Intersection over Union (IoU), Accuracy (Acc), and Dice Coefficient (Dice), all reported as percentages (\%). Furthermore, we provided the proposed method's trainable parameters and their proportion relative to the frozen VFM backbone, denoted as ``*Params". The IoU quantifies the overlap between the predicted and the ground truth masks:

\begin{equation}
\text{IoU}_i = \frac{| \text{Pred}_i \cap \text{GT}_i |}{| \text{Pred}_i \cup \text{GT}_i |} = \frac{TP_i}{TP_i + FP_i + FN_i},
\end{equation}
where \( TP_i \), \( FP_i \), and \( FN_i \) represent the true positives, false positives, and false negatives for class \( i \), respectively. The Acc represents the pixel-wise classification accuracy:
\begin{equation}
    Acc_i = \frac{TP_i + TN_i}{TP_i + TN_i + FP_i + FN_i},
\end{equation}
where $TN_i$ represents the true negatives for class $i$. The Dice measures the similarity between the predicted and ground truth masks, and it is equivalent to the F1-score metric:
\begin{equation}
    Dice_i = \frac{2 | \text{Pred}_i \cap \text{GT}_i |}{| \text{Pred}_i | + | \text{GT}_i |} = \frac{2TP_i}{2TP_i + FP_i + FN_i}.
\end{equation}

The metrics mAcc, mIoU, and mDice represent the mean values computed for each class, offering insight into the model’s performance at the class level. In contrast, the overall accuracy (aAcc) evaluates the model's classification performance across the entire dataset. These metrics are all positive indicators, with higher values signifying better performance, simplifying interpretation, and facilitating model comparison.


\subsection{Ablation Studies and Analysis}\label{sec: ablation}

\noindent{This section presents a series of comprehensive ablation studies to evaluate the effectiveness of key components in our proposed model using the CloudSEN12\_High\_L1C dataset.}

\subsubsection{Variants of VFM's Backbone}
\begin{table}[ht]
\setlength{\tabcolsep}{1.25mm}
\centering
\caption{Cloud segmentation performance comparison of different VFM's backbone with their multiple variants.}
\label{tab: backbone}
\begin{tabular}{c|c|cccc|c}
\toprule
\textbf{Backbone} & \textbf{Variant} & \textbf{mIoU} & \textbf{mAcc} & \textbf{aAcc} & \textbf{mDice} & \textbf{*Params (M)} \\ \midrule
& Base & 72.53 & 83.29 & 89.46 & 83.22 & 0.92 (1.1\%) \\
& Large & 73.36 & 83.90 & 89.84 & 83.85 & 1.83 (0.6\%) \\
\multirow{-3}{*}{SAM~\cite{sam}} & \cellcolor[HTML]{D9D9D9}Huge & \cellcolor[HTML]{D9D9D9}73.70 & \cellcolor[HTML]{D9D9D9}84.37 & \cellcolor[HTML]{D9D9D9}89.95 & \cellcolor[HTML]{D9D9D9}84.07 & \cellcolor[HTML]{D9D9D9}2.63 (\textbf{0.4\%}) \\ \midrule
& Small & 68.12 & 79.37 & 87.50 & 79.84 & \textbf{0.77} (3.4\%) \\
& Base & 71.95 & 83.02 & 89.22 & 82.82 & 0.92 (1.1\%) \\
\multirow{-3}{*}{\textbf{DINOv2}~\cite{dinov2}} & \cellcolor[HTML]{D9D9D9}\textbf{Large} & \cellcolor[HTML]{D9D9D9}\textbf{74.18} & \cellcolor[HTML]{D9D9D9}\textbf{84.79} & \cellcolor[HTML]{D9D9D9}\textbf{90.19} & \cellcolor[HTML]{D9D9D9}\textbf{84.46} & \cellcolor[HTML]{D9D9D9}1.82~(0.6\%) \\ \bottomrule
\end{tabular}
\end{table}

To evaluate the impact of different vision foundation models and their variants on our method's performance, we conducted comprehensive experiments using SAM~\cite{sam} and DINOv2~\cite{dinov2} as backbones. Table \ref{tab: backbone} compares three available model variants.

We observed that SAM and DINOv2 exhibited consistent performance improvements as the model scale increased. Among the SAM variants, the Huge version performed best, achieving 73.70\% mIoU, 84.37\% mAcc, 89.95\% aAcc, and 84.07\% mDice. In contrast, DINOv2 (Large) demonstrated superior performance across all metrics, reaching 74.18\% mIoU, 84.79\% mAcc, 90.19\% aAcc, and 84.46\% mDice.

More importantly, DINOv2 (Large) was significantly more computationally efficient, requiring only 11 hours and 8 minutes of training time and 14,558M of GPU memory. In comparison, the SAM (Huge) variant took 19 hours and 9 minutes to train, with GPU memory usage of 24,892M, which was substantially higher than DINOv2 (Large).

These results indicate that DINOv2 (Large) outperforms SAM (Huge) across all metrics while being significantly more efficient regarding training time and memory usage. Therefore, DINOv2 (Large) emerges as the optimal backbone for our method, striking an excellent balance between performance and computational efficiency for cloud segmentation.

\subsubsection{ConvNet Architectures of SPM}
\begin{table}[ht]
\centering
\caption{Comparison of cloud segmentation performance and model parameters for different ConvNet architectures.}
\label{tab: block}
\setlength{\tabcolsep}{1.5mm}
\begin{tabular}{c|cccc|c}
\toprule
\textbf{ConvNet Block}          & \textbf{mIoU}  & \textbf{mAcc}  & \textbf{aAcc}  & \textbf{mDice} & \textbf{*Params (M)}  \\ \midrule
Transformer-Like~\cite{convnext} & 74.08 & \textbf{84.95} & 90.09 & 84.39 & \textbf{1.81}~(\textbf{0.6}\%) \\
\rowcolor{gray!30}\textbf{Pure-ConvNet}~\cite{pmaa}         & \textbf{74.18} & 84.79 & \textbf{90.19} & \textbf{84.46} & 1.82~(0.6\%) \\ \bottomrule
\end{tabular}
\end{table}
We investigate the impact of different ConvNet architectures on our method's performance, as shown in Table \ref{tab: block}. Specifically, we compare two types of ConvNet designs: a Transformer-Like architecture inspired by~\cite{convnext} and a Pure-ConvNet structure from~\cite{pmaa}.

The Pure-ConvNet~\cite{pmaa} architecture achieves slightly better overall performance with 74.18\% mIoU, 90.19\% aAcc, and 84.46\% mDice, while the Transformer-Like~\cite{convnext} design shows marginally higher mAcc at 84.95\%. Both architectures maintain similar parameter efficiency, with approximately 1.8M.

The comparable performance between these two architectures suggests that both designs are viable options for our method. However, considering the slightly higher performance in most metrics and the simpler structure, we adopt the Pure-ConvNet architecture as our default choice.

\subsubsection{Dimension of Interactive Context}
\begin{figure}[ht]
\centering
\includegraphics[width=\linewidth]{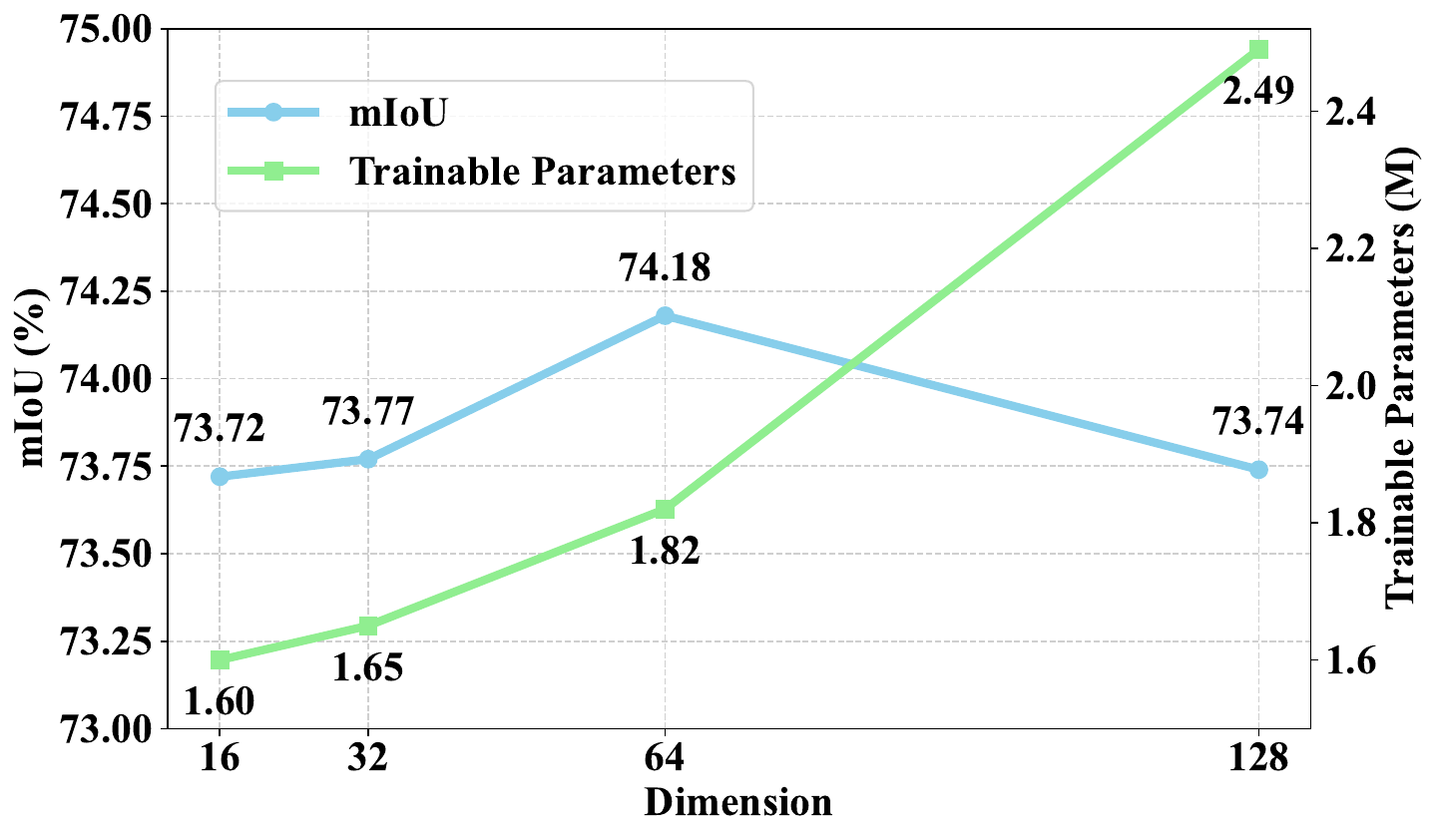}
\caption{Ablation study of different dimension settings on the interactive context extracted by the spatial perception module.}
\label{fig: DimComparison}
\end{figure}

We conduct experiments to explore the impact of different dimension settings on the interactive context extracted by the SPM. As illustrated in Fig.~\ref{fig: DimComparison}, we evaluate four configurations: 16, 32, 64, and 128.

The results show that the model's performance does not necessarily improve with larger dimensions. Starting from dimension 16 with a mIoU of 73.72\%, we observe a slight improvement to 73.77\% when increasing to dimension 32. The peak performance is achieved at dimension 64 with a mIoU of 74.18\%. However, further increasing the dimension to 128 leads to slight performance degradation (0.44\% dropping).

Meanwhile, the number of trainable parameters grows steadily with the dimension size. The model requires 1.60M parameters at dimension 16, gradually increasing to 1.65M at dimension 32, 1.82M at dimension 64, and significantly jumping to 2.49M at dimension 128. Based on these observations, we identify dimension 64 as the optimal choice, offering the best trade-off between performance and efficiency.

\begin{figure}[ht]
\centering
\includegraphics[width=\linewidth]{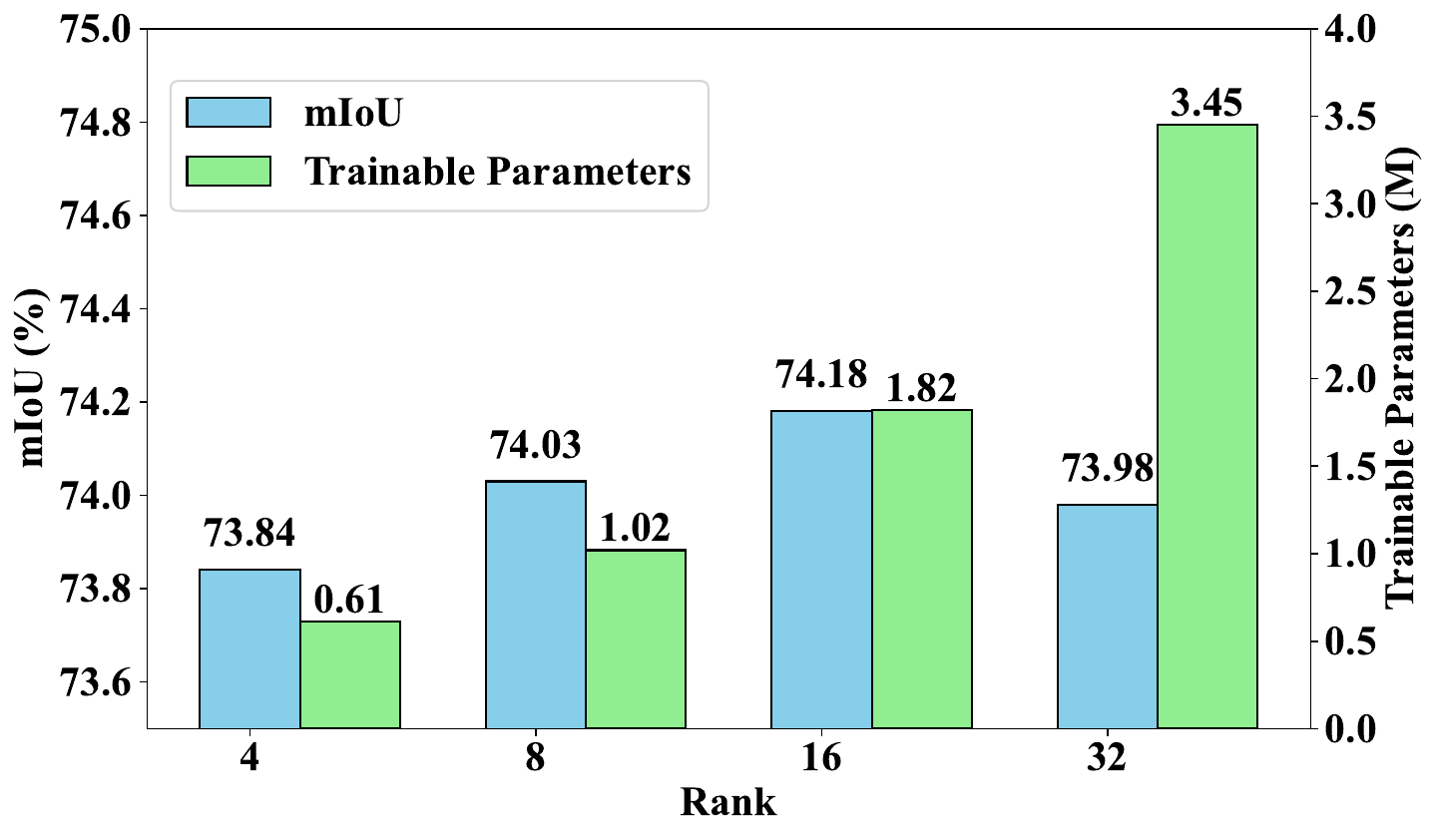}
\caption{Study of the low-rank MLP in the adapting module.}
\label{fig: RankBar}
\end{figure}

\subsubsection{Low-Rank MLP of Adapting Module}\label{sec: condition}

We investigate the influence of different rank settings on our model's performance and parameter efficiency. The rank dimension controls the size of the low-rank matrices used to approximate the weight updates in the MLP of the adapting module, which helps to reduce the number of trainable parameters. As shown in Figure~\ref{fig: RankBar}, we examine four rank settings: 4, 8, 16, and 32.

The experimental results reveal an interesting relationship between rank size and model performance. With a rank of 4, the model achieves a mIoU of 73.84\% while requiring only 0.61M trainable parameters. Increasing the rank to 8 improves performance to 74.03\% with 1.02M parameters. The peak performance is reached at rank 16 with a mIoU of 74.18\%, which requires 1.82M parameters—further increasing the rank to 32 leads to a slight performance degradation. Based on these observations, we choose rank 16 as our optimal setting. This finding aligns with LoRA's design principle~\cite{lora} of achieving efficient adaptation through low-rank approximation.

\subsubsection{Importance of the Aggregator}
\begin{table}[ht]
\centering
\setlength{\tabcolsep}{2.25mm}
\caption{Cloud segmentation performance comparison of context output strategies for the spatial perception module.}
\label{tab: aggregator}
\begin{tabular}{c|cccc|c}
\toprule
\textbf{Feature Type}      & \textbf{mIoU}  & \textbf{mAcc}  & \textbf{aAcc}  & \textbf{mDice} & \textbf{*Params (M)}  \\ \midrule
Single-Scale & 74.05 & 84.45 & 90.19 & 84.34 & 1.82~(0.6\%) \\
Multi-Scale  & 74.10 & 84.70 & 90.15 & 84.40 & 1.82~(0.6\%) \\
\rowcolor{gray!30}\textbf{Aggregated}   & \textbf{74.18} & \textbf{84.79} & \textbf{90.19} & \textbf{84.46} & 1.82~(0.6\%) \\ \bottomrule
\end{tabular}
\end{table}
We explore three context output strategies for our SPM: Single-Scale, Multi-Scale, and Aggregated approaches. Table \ref{tab: aggregator} presents the comparative results of these strategies while maintaining the same parameter count of 1.82M (0.6\% of total parameters).

The ``Single-Scale" strategy, which directly utilizes the final feature output from the last ConvNet block, achieves baseline performance with 74.05\% mIoU, 84.45\% mAcc, and 84.34\% mDice. The ``Multi-Scale" approach, leveraging multiple scale features without fusion, shows slight improvements with 74.10\% mIoU, 84.70\% mAcc, and 84.40\% mDice.

The ``Aggregated" strategy, which fuses multiple scale features into a single output with parameter-free operation, demonstrates the best overall performance. It achieves 74.18\% mIoU, 84.79\% mAcc, and 84.46\% mDice while maintaining the same aAcc (90.19\%) as the ``Single-Scale" approach. This superior performance can be attributed to the strategy's ability to combine and utilize information from different feature scales effectively. Based on these results, we adopt the ``Aggregated" feature output strategy in our final model design, as it provides the most effective use of dense multi-scale feature information without introducing additional parameters.

\subsubsection{Interaction Frequency of Adapting Module}\label{sec: transformer_ablation}
\begin{table}[ht]
\centering
\caption{Comparison of performance at different interaction frequencies ($N$) between the Adapter and Transformer layers.}
\label{tab: interaction}
\setlength{\tabcolsep}{3.4mm}
\begin{tabular}{c|cccc|c}
\toprule
\textbf{$N$} & \textbf{mIoU}  & \textbf{mAcc}  & \textbf{aAcc}  & \textbf{mDice} & \textbf{*Params (M)}  \\ \midrule
4     & 73.79 & 84.22 & 90.10 & 84.17 & \textbf{0.47}~(\textbf{0.2}\%) \\
8   & 73.76 & 84.46 & 89.99 & 84.15 & 0.74~(0.2\%) \\
12   & 73.75 & 84.46 & 89.99 & 84.13 & 1.02~(0.3\%) \\
\rowcolor{gray!30}\textbf{24}   & \textbf{74.18} & \textbf{84.79} & \textbf{90.19} & \textbf{84.46} & 1.82~(0.6\%) \\ \bottomrule
\end{tabular}
\end{table}
In this ablation study, we explore the effect of varying the interaction frequency ($N$) between the Adapter and Transformer layers on the model's performance. The parameter $N$ indicates how frequently the Adapter interacts with the Transformer layers, with larger values corresponding to more frequent interactions. Table \ref{tab: interaction} presents the results of this experiment, evaluating different values of $N$: 4, 8, 12, and 24. Specifically, when $N$ is set to 4, the Adapter interacts with the Transformer layers at indices [0, 6, 12, 18]. The ``8" and ``12" represent interactions at every 3rd and 2nd layer starting from index 0, respectively.

The results indicate that the optimal performance is achieved when $N$ is set to 24, where the model attains 74.18\% mIoU, 84.79\% mAcc, 90.19\% aAcc, and 84.46\% mDice. This configuration outperforms all others, particularly in terms of mIoU and mAcc. Notably, the increase in interaction frequency from 4 to 24 improves the model's ability to capture complex spatial dependencies, significantly boosting the performance metrics.

In contrast, when the interaction frequency is lower ($N$ = 4, 8, or 12), performance remains relatively stable but does not reach the peak achieved by the 24-interaction configuration. Furthermore, while the number of parameters increases with higher values of $N$, the increase is minimal (only 0.2-0.6\% in parameters), suggesting that the additional interactions do not incur significant computational overhead, thanks to the parameter-efficient LoRA-MLP design. These results suggest that allowing the Adapter to interact with more Transformer layers improves model performance, with the 24-interaction configuration representing a sweet spot for this balance.

\subsubsection{Overall Components in the Cloud-Adapter}\label{sec: architecture}
\begin{table}[ht]
\centering
\setlength{\tabcolsep}{0.3mm}
\caption{Study of each component in the Cloud-Adapter.}
\label{tab: existing}
\begin{tabular}{ccc|c|c|c}
\toprule
\multicolumn{3}{c|}{\textbf{Spatial Perception Module}} & \multirow{2}{*}{\textbf{Adapting Module}} & \multirow{2}{*}{\textbf{mIoU}} & \multirow{2}{*}{\textbf{*Params (M})} \\ \cline{1-3}
\textbf{Stem}     & \textbf{ConvNet Block}     & \textbf{Aggregator}     &                                  &                       &                             \\ \midrule 
\(\times\)        & \(\times\)                 & \(\times\)              & \(\times\)                                & 73.10                 & \textbf{0.00}               \\
\(\checkmark\)        & \(\times\)                 & \(\times\)              & \(\times\)                                & 73.12                 & 3e-4                     \\
\(\checkmark\)        & \(\checkmark\)                 & \(\times\)              & \(\times\)                                & 73.20                 & 0.20                        \\
\(\checkmark\)        & \(\checkmark\)                 & \(\checkmark\)              & \(\times\)                                & 73.37                 & 0.20                        \\
\rowcolor{gray!30} \(\checkmark\)        & \(\checkmark\)                 & \(\checkmark\)              & \(\checkmark\)                                & \textbf{74.18}        & 1.82   \\ \bottomrule                      
\end{tabular}
\end{table}
To evaluate the contribution of each component in our model, we conducted an ablation study by systematically removing or modifying specific modules. The results are presented in Table \ref{tab: existing}, showing the performance in terms of mIoU and parameter count for different configurations. Notably, in configurations without the adapting module, the features output by the spatial perception module are concatenated with the features from the VFM backbone before being passed to the head for prediction.

In the baseline model, without any components from the spatial perception module or adapting module, the features from the VFM backbone were directly passed to the head for segmentation prediction. This configuration achieved a mIoU of 73.10\%. When the stem module was added, the mIoU increased slightly to 73.12\%, with a negligible parameter increase ($3\times10^{-4}$M). Including the stem module helps in early spatial feature projection and extraction, providing the model with rudimentary spatial awareness information.

Further improvement was seen when the ConvNet Block and the stem module were added to the model. This resulted in a mIoU of 73.20\% and added 0.20M parameters. The ConvNet Block enhances the model's ability to extract more complex spatial features, contributing to a slight performance boost. When the aggregator module was incorporated, the mIoU rose to 73.37\%, with no additional parameters, indicating that the aggregator effectively combines multi-scale features to improve the model’s overall cloud segmentation performance.

Finally, with the inclusion of the adapting module, the full model achieved the highest mIoU of 74.18\% and a total of 1.82M parameters. The adapting module enables fine-tuned integration of the spatial and backbone features, significantly enhancing the model’s performance despite increased parameters. This demonstrates the crucial role of the adapting module in optimizing the feature fusion process. In conclusion, the ablation study confirms that each component contributes incrementally to the model's performance. The concatenation of spatial and backbone features and successive refinement and adaptation through the added modules substantially improved the segmentation performance in particular domains.

\subsection{Comparison with State-of-the-Art Methods}

\noindent{We compare the performance of our proposed Cloud-Adapter with several state-of-the-art cloud segmentation methods across different datasets: HRC\_WHU, GF12MS\_WHU\_GF1, GF12MS\_WHU\_GF2, CloudSEN12\_High\_L1C, and CloudSEN12\_High\_L2A. These datasets include both binary (HRC\_WHU, GF1, GF2) and multi-class (L1C, L2A) cloud segmentation. The performance comparison, as shown in Tables~\ref{tab: hrc_whu}, \ref{tab: gf12ms_whu_gf1}, \ref{tab: gf12ms_whu_gf2}, \ref{tab: cloudsen12_high_l1c}, and \ref{tab: cloudsen12_high_l2a}, highlights Cloud-Adapter's robustness in both binary and multi-class cloud segmentation.

\subsubsection{Binary-Class Cloud Segmentation}

\begin{table}[!t]
    \centering
    \setlength{\tabcolsep}{3.2mm}
    \caption{Cloud segmentation performance comparison of different methods on the HRC\_WHU dataset.}
    \label{tab: hrc_whu}
    \begin{tabular}{l|cccccc}
        \toprule
        \textbf{Method} & \textbf{aAcc} & \textbf{mIoU} & \textbf{mAcc} & \textbf{mDice} \\ \midrule
        MCDNet~\cite{mcdnet}     & 75.14 & 53.50 & 68.91 & 67.96 \\ 
        SCNN~\cite{scnn}       & 74.51 & 57.22 & 81.27 & 72.31 \\ 
        KappaMask~\cite{kappamask}  & 84.73 & 67.48 & 80.30 & 79.74 \\
        CDNetv2~\cite{cdnetv2}    & 89.71 & 76.75 & 87.46 & 86.46 \\ 
        DBNet~\cite{dbnet}      & 90.11 & 77.78 & 88.80 & 87.17 \\ 
        CDNetv1~\cite{cdnetv1}    & 89.88 & 77.79 & 89.93 & 87.20 \\
        RSAM-Seg~\cite{rsam-seg}    & 92.07 & 80.90 & 88.72 & 89.16 \\
        UNetMobv2~\cite{cloudsen12_high}  & 92.13 & 79.91 & 85.61 & 88.45 \\
        HRCloudNet~\cite{hrcloudnet} & 92.93 & 83.44 & 92.39 & 90.79 \\ 
        \rowcolor{gray!30}\textbf{Cloud-Adapter~(Ours)} &\textbf{94.50} &\textbf{89.05} &\textbf{93.74} &\textbf{94.19} \\
        \bottomrule
    \end{tabular}
\end{table}

\begin{table}[!t]
    \centering
    \caption{Cloud segmentation performance comparison of different methods on the GF12MS\_WHU\_GF1 dataset.}
    \label{tab: gf12ms_whu_gf1}
    \setlength{\tabcolsep}{3.3mm}
    \begin{tabular}{l|cccccc}
        \toprule
        \textbf{Method} & \textbf{aAcc} & \textbf{mIoU} & \textbf{mAcc} & \textbf{mDice} \\ \midrule
        SCNN~\cite{scnn}         & 97.18 & 81.68 & 87.21 & 89.13 \\
        CDNetv1~\cite{cdnetv1}      & 96.81 & 81.82 & 92.75 & 89.27 \\
        CDNetv2~\cite{cdnetv2}      & 97.55 & 84.93 & 92.96 & 91.36 \\
        MCDNet~\cite{mcdnet}       & 97.55 & 85.16 & 93.97 & 91.51 \\
        RSAM-Seg~\cite{rsam-seg}   & 98.36 & 89.26 & 94.70 & 94.09 \\ 
        UNetMobv2~\cite{cloudsen12_high}    & 98.82 & 91.71 & 93.99 & 95.53 \\
        DBNet~\cite{dbnet}        & 98.73 & 91.36 & 95.19 & 95.33 \\
        HRCloudNet~\cite{hrcloudnet}   & 98.80 & 91.86 & 95.82 & 95.62 \\
        KappaMask~\cite{kappamask}    & 98.91 & 92.42 & 95.05 & 95.94 \\
        \rowcolor{gray!30}\textbf{Cloud-Adapter~(Ours)} &\textbf{98.92} &\textbf{92.55} &\textbf{95.99} &\textbf{96.02} \\
        \bottomrule
    \end{tabular}
\end{table}

\begin{table}[!t]
    \centering
    \caption{Cloud segmentation performance comparison of different methods on the GF12MS\_WHU\_GF2 dataset.}
    \label{tab: gf12ms_whu_gf2}
    \setlength{\tabcolsep}{3.2mm}
    \begin{tabular}{l|cccccc}
        \toprule
        \textbf{Method} & \textbf{aAcc} & \textbf{mIoU} & \textbf{mAcc} & \textbf{mDice} \\ \midrule
        KappaMask~\cite{kappamask}    & 90.30  & 72.00  & 77.64 & 82.57 \\
        HRCloudNet~\cite{hrcloudnet}  & 91.46 & 75.57 & 80.95 & 85.29 \\
        SCNN~\cite{scnn}             & 91.99 & 76.99 & 82.06 & 86.32 \\
        CDNetv1~\cite{cdnetv1}        & 92.42 & 78.20  & 83.07 & 87.17 \\
        MCDNet~\cite{mcdnet}         & 92.30  & 78.36 & 83.95 & 87.31 \\
        DBNet~\cite{dbnet}           & 92.61 & 78.68 & 83.39 & 87.50  \\
        CDNetv2~\cite{cdnetv2}       & 92.63 & 78.84 & 83.67 & 87.61 \\
        UNetMobv2~\cite{cloudsen12_high} & 93.22 & 80.44 & 84.86 & 88.70  \\
        \rowcolor{gray!30}\textbf{Cloud-Adapter~(Ours)} &\textbf{94.04} &\textbf{83.02} &\textbf{87.54} &\textbf{90.40} \\
        \bottomrule
    \end{tabular}
\end{table}

\begin{figure*}[!t]
\centering
\includegraphics[width=0.95\linewidth]{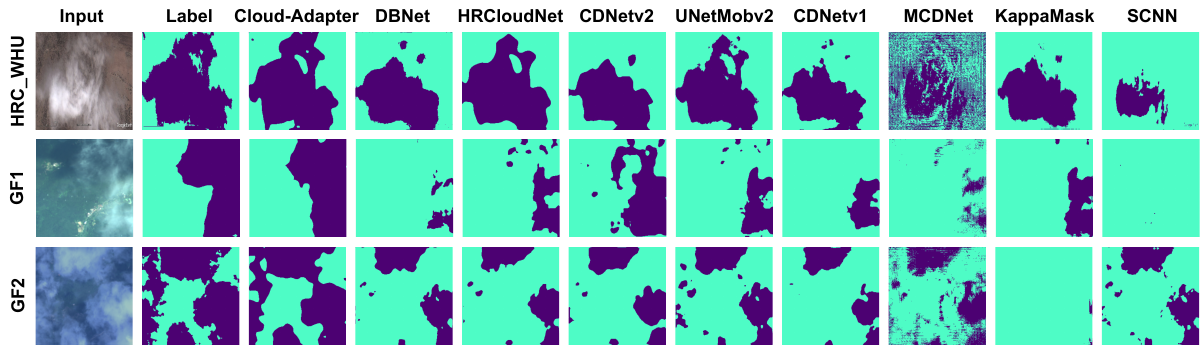}
\begin{tikzpicture}
    \node {
        \begin{tabular}{cccc}
            \tikz\fill[clear sky] (0,0) rectangle (0.3,0.3); & Clear Sky &
            \tikz\fill[thick cloud] (0,0) rectangle (0.3,0.3); & Thick Cloud \\
        \end{tabular}
    };
\end{tikzpicture}
\vspace{-2ex}
\caption{Comparison of visualized segmentation results of different models on the coarse-grained cloud segmentation dataset.}
\label{fig: hrc_whu_gf1_gf2}
\end{figure*}
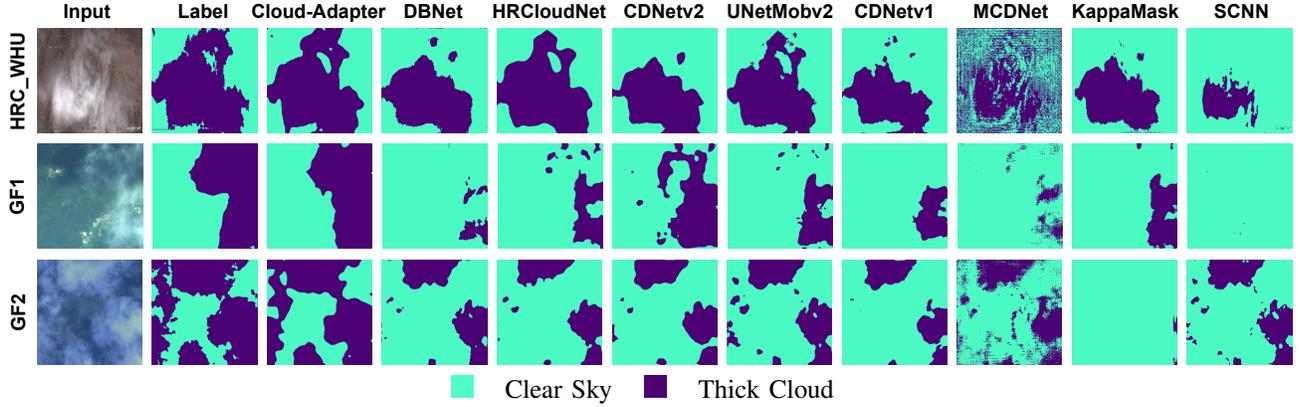

\paragraph{Quantitative Comparison}

Cloud-Adapter consistently demonstrates superior performance across all evaluation metrics (aAcc, mIoU, mAcc, and mDice) for binary-class cloud segmentation datasets. On the HRC\_WHU dataset (Table~\ref{tab: hrc_whu}), Cloud-Adapter achieves the highest scores with an aAcc of 94.50\%, mIoU of 89.05\%, mAcc of 93.74\%, and mDice of 94.19\%, significantly outperforming the second-best method, HRCloudNet, which only achieves a mIoU of 83.44\%. On the GF12MS\_WHU\_GF1 dataset (Table~\ref{tab: gf12ms_whu_gf1}), Cloud-Adapter attains an aAcc of 98.92\%, mIoU of 92.55\%, mAcc of 95.99\%, and mDice of 96.02\%, surpassing KappaMask, which achieved a slightly lower mIoU of 92.42\%. On the GF12MS\_WHU\_GF2 dataset (Table~\ref{tab: gf12ms_whu_gf2}), Cloud-Adapter maintains its leading performance with an aAcc of 94.04\%, mIoU of 83.02\%, mAcc of 87.54\%, and mDice of 90.40\%, showing its robustness in handling challenging satellite data.

\paragraph{Visualization Comparison}

Fig.~\ref{fig: hrc_whu_gf1_gf2} compares the visualized segmentation results for three binary-class datasets. Cloud-Adapter stands out for its ability to accurately capture cloud boundaries and reduce over-segmentation issues, which is common in other methods like DBNet and HRCloudNet. The results illustrate its superiority in segmenting clear sky and cloud regions, even under complex conditions.

\subsubsection{Multi-Class Cloud Segmentation}

\begin{table*}[!t]
    \centering
    \caption{Cloud segmentation performance comparison of different methods on 
 the CloudSEN12\_High\_L1C dataset.}
    \label{tab: cloudsen12_high_l1c}
    \setlength{\tabcolsep}{1.3mm}
\begin{tabular}{l|llll|c|llll|l|llll|l|l}
\toprule
\multirow{2}{*}{\textbf{Method}} & \multicolumn{4}{c|}{\textbf{IoU}} & \multirow{2}{*}{\textbf{mIoU}} & \multicolumn{4}{c|}{\textbf{Acc}} & \multirow{2}{*}{\textbf{mAcc}} & \multicolumn{4}{c|}{\textbf{Dice}} & \multirow{2}{*}{\textbf{mDice}} & \multirow{2}{*}{\textbf{aAcc}} \\ \cline{2-5} \cline{7-10} \cline{12-15}
                        & CRS  & TNC  & TKC  & CDS &                       & CRS  & TNC  & TKC  & CDS &                       & CRS  & TNC  & TKC  & CDS  &                        &          \\ \midrule 
SCNN~\cite{scnn}        & 56.39 & 00.00  & 34.60 & 00.00  & 22.75 & 89.69 & 00.00 & 45.08 & 00.00  & 33.69 & 72.11 & 00.00 & 51.41 & 00.00  & 30.88 & 60.19 \\  
KappaMask~\cite{kappamask}   & 76.84 & 14.75 & 73.51 & 00.00  & 41.27 & \textbf{97.54} & 27.78 & 77.73 & 00.00  & 50.76 & 86.91 & 25.70 & 84.73 & 00.00  & 49.33 & 76.27 \\  
MCDNet~\cite{mcdnet}      & 68.39 & 13.02 & 65.26 & 32.52 & 44.80 & 77.85 & 18.39 & 84.49 & 59.20 & 59.99 & 81.23 & 23.04 & 78.98 & 49.08 & 58.08 & 72.68 \\  
CDNetv1~\cite{cdnetv1}    & 80.48 & 36.50 & 79.60 & 44.83 & 60.35 & 90.60 & 50.19 & 88.34 & 60.72 & 72.46 & 89.19 & 53.48 & 88.64 & 61.91 & 73.30 & 83.48 \\  
DBNet~\cite{dbnet}        & 85.20 & 43.02 & 81.40 & 52.44 & 65.52 & 95.56 & 53.15 & 89.40 & 62.47 & 75.15 & 92.01 & 60.16 & 89.75 & 68.80 & 77.68 & 86.83 \\  
CDNetv2~\cite{cdnetv2}    & 84.98 & 43.37 & 80.67 & 53.40 & 65.60 & 94.80 & 53.13 & 89.32 & 65.34 & 75.65 & 91.88 & 60.50 & 89.30 & 69.62 & 77.83 & 86.68 \\  
HRCloudNet~\cite{hrcloudnet} & 86.01 & 45.88 & 83.54 & 57.61 & 68.26 & 94.49 & 59.47 & 91.39 & 67.19 & 78.13 & 92.48 & 62.90 & 91.03 & 73.10 & 79.88 & 87.86 \\  
UNetMobv2~\cite{cloudsen12_high}  & 88.51 & 50.12 & 84.79 & 63.17 & 71.65 & 95.77 & 61.87 & 91.44 & 75.13 & 81.05 & 93.90 & 66.78 & 91.77 & 77.43 & 82.47 & 89.52 \\  
\rowcolor{gray!30}\textbf{Cloud-Adapter~(Ours)} & \textbf{89.19} & \textbf{56.15} & \textbf{85.46} & \textbf{65.93} & \textbf{74.18} & 94.08 & \textbf{75.23} & \textbf{91.72} & \textbf{78.15} & \textbf{84.79} & \textbf{94.29} & \textbf{71.91} & \textbf{92.16} & \textbf{79.47} & \textbf{84.46} & \textbf{90.19} \\  
\bottomrule
\end{tabular}
\end{table*}

\begin{table*}[!t]
    \centering
    \caption{Cloud segmentation performance comparison of different methods on the CloudSEN12\_High\_L2A dataset.}
    \label{tab: cloudsen12_high_l2a}
    \setlength{\tabcolsep}{1.3mm}
\begin{tabular}{l|llll|c|llll|c|llll|c|l}
\toprule
\multirow{2}{*}{\textbf{Method}} & \multicolumn{4}{c|}{\textbf{IoU}} & \multirow{2}{*}{\textbf{mIoU}} & \multicolumn{4}{c|}{\textbf{Acc}} & \multirow{2}{*}{\textbf{mAcc}} & \multicolumn{4}{c|}{\textbf{Dice}} & \multirow{2}{*}{\textbf{mDice}} & \multirow{2}{*}{\textbf{aAcc}} \\ \cline{2-5} \cline{7-10} \cline{12-15}
                        & CRS  & TNC  & TKC  & CDS &                       & CRS  & TNC  & TKC  & CDS &                       & CRS  & TNC  & TKC  & CDS  &                        &          \\ \midrule 
SCNN~\cite{scnn}        & 62.20 & 00.00  & 52.74 & 00.08  & 28.76 & 85.73 & 00.00 & 76.19 & 00.08  & 40.50 & 76.70 & 00.00 & 69.06 & 00.16  & 36.48 & 67.14 \\  
KappaMask~\cite{kappamask}   & 79.58 & 23.35 & 78.20 & 00.00  & 45.28 & \textbf{97.24} & 38.05 & 86.58 & 00.00  & 55.47 & 88.63 & 37.85 & 87.77 & 00.00  & 53.56 & 79.60 \\  
MCDNet~\cite{mcdnet}      & 72.46 & 14.96 & 67.04 & 31.62 & 46.52 & 85.76 & 21.09 & 85.29 & 42.01 & 58.54 & 84.03 & 26.02 & 80.27 & 48.04 & 59.59 & 75.67 \\  
CDNetv1~\cite{cdnetv1}    & 82.58 & 38.61 & 78.72 & 49.64 & 62.39 & 92.73 & 52.65 & 88.12 & 60.50 & 73.50 & 90.46 & 55.71 & 88.09 & 66.35 & 75.15 & 84.74 \\  
DBNet~\cite{dbnet}        & 85.42 & 42.76 & 80.80 & 53.62 & 65.65 & 94.83 & 53.53 & 90.27 & 63.45 & 75.52 & 92.14 & 59.90 & 89.38 & 69.81 & 77.81 & 86.82 \\  
CDNetv2~\cite{cdnetv2}    & 85.35 & 44.01 & 80.85 & 54.00 & 66.05 & 95.65 & 56.37 & 88.06 & 63.56 & 75.91 & 92.09 & 61.12 & 89.41 & 70.13 & 78.19 & 86.88 \\  
HRCloudNet~\cite{hrcloudnet} & 87.24 & 44.18 & 82.78 & 59.20 & 68.35 & 95.41 & 51.00 & 93.17 & 69.83 & 77.35 & 93.19 & 61.28 & 90.58 & 74.37 & 79.85 & 88.35 \\  
UNetMobv2~\cite{cloudsen12_high}  & 87.76 & 47.38 & 84.16 & 62.13 & 70.36 & 95.98 & 59.60 & 90.94 & 71.75 & 79.57 & 93.48 & 64.29 & 91.40 & 76.64 & 81.45 & 88.96 \\  
\rowcolor{gray!30}\textbf{Cloud-Adapter~(Ours)} & \textbf{89.04} & \textbf{54.97} & \textbf{84.91} & \textbf{64.60} & \textbf{73.38} & 94.65 & \textbf{73.64} & \textbf{90.61} & \textbf{76.81} & \textbf{83.93} & \textbf{94.20} & \textbf{70.95} & \textbf{91.84} & \textbf{78.49} & \textbf{83.87} & \textbf{89.90} \\  
\bottomrule
\end{tabular}
\end{table*}

\begin{table*}[!t]
\centering
\caption{Cloud segmentation performance comparison of different methods on the L8\_Biome dataset.}
    \label{tab: l8_biome}
    \setlength{\tabcolsep}{1.3mm}
\begin{tabular}{l|llll|c|llll|c|llll|c|l}
\toprule
\multicolumn{1}{l|}{\multirow{2}{*}{\textbf{Method}}} & \multicolumn{4}{c|}{\textbf{IoU}}                                           & \multicolumn{1}{c|}{\multirow{2}{*}{\textbf{mIoU}}} & \multicolumn{4}{c|}{\textbf{Acc}}                                           & \multicolumn{1}{c|}{\multirow{2}{*}{\textbf{mAcc}}} & \multicolumn{4}{c|}{\textbf{Dice}}                                          & \multicolumn{1}{c|}{\multirow{2}{*}{\textbf{mDice}}} & \multirow{2}{*}{\textbf{aAcc}} \\ \cline{2-5} \cline{7-10} \cline{12-15}
\multicolumn{1}{l|}{}                                 & CRS            & TNC            & TKC            & \multicolumn{1}{l|}{CDS} & \multicolumn{1}{c|}{}                               & CRS            & TNC            & TKC            & \multicolumn{1}{l|}{CDS} & \multicolumn{1}{c|}{}                               & CRS            & TNC            & TKC            & \multicolumn{1}{l|}{CDS} & \multicolumn{1}{c|}{}                                &                                \\ \midrule
MCDNet~\cite{mcdnet}                                                & 65.83          & 08.40          & 60.97          & 00.21                    & 33.85                                               & 73.80          & 10.26          & 95.01          & 00.22                    & 44.82                                               & 79.40          & 15.49          & 75.75          & 00.41                    & 42.76                                                & 69.75                          \\
SCNN~\cite{scnn}                                                  & 66.30          & 00.00          & 63.24          & 00.00                    & 32.38                                               & 81.11          & 00.00          & \textbf{95.12} & 00.00                    & 44.06                                               & 79.73          & 00.00          & 77.48          & 00.00                    & 39.30                                                & 71.22                          \\
CDNetv1~\cite{cdnetv1}                                               & 60.69          & 19.63          & 56.79          & 01.21                    & 34.58                                               & 88.19          & 25.51          & 67.43          & 01.22                    & 45.59                                               & 75.54          & 32.81          & 72.44          & 02.40                    & 45.80                                                & 68.16                          \\
KappaMask~\cite{kappamask}                                             & 77.72          & 22.63          & 68.12          & 00.00                    & 42.12                                               & 93.08          & 32.87          & 80.96          & 00.00                    & 51.73                                               & 87.45          & 36.91          & 81.04          & 00.00                    & 51.35                                                & 76.63                          \\
UNetMobv2~\cite{cloudsen12_high}                                             & 79.18          & 28.35          & 80.46          & 03.05                    & 47.76                                               & 93.62          & 34.63          & 93.59          & 03.31                    & 56.29                                               & 88.37          & 44.18          & 89.16          & 05.92                    & 56.91                                                & 82.00                          \\
CDNetv2~\cite{cdnetv2}                                               & 73.81          & 24.85          & 75.85          & 00.00                    & 43.63                                               & 90.18          & 31.95          & 89.81          & 00.00                    & 52.98                                               & 84.93          & 39.81          & 86.27          & 00.00                    & 52.75                                                & 78.56                          \\
HRCloudNet~\cite{hrcloudnet}                                            & 72.41          & 30.89          & 70.76          & 00.00                    & 43.51                                               & 85.50          & 43.70          & 85.88          & 00.00                    & 53.77                                               & 83.99          & 47.20          & 82.88          & 00.00                    & 53.52                                                & 77.04                          \\
DBNet~\cite{dbnet}                                                 & 80.19          & 38.18          & \textbf{82.37} & 04.90                    & 51.41                                               & \textbf{96.51} & 49.34          & 87.89          & 05.03                    & 59.70                                               & 89.00          & 55.25          & \textbf{90.34} & 09.34                    & 60.99                                                & \textbf{83.62}                 \\
\rowcolor{gray!30}\textbf{Cloud-Adapter~(Ours)}                                   & \textbf{85.81} & \textbf{44.41} & 70.75          & \textbf{29.13}           & \textbf{57.53}                                      & 90.63          & \textbf{78.76} & 75.35          & \textbf{32.65}           & \textbf{69.34}                                      & \textbf{92.36} & \textbf{61.51} & 82.87          & \textbf{45.12}           & \textbf{70.46}                                       & 81.79            \\ \bottomrule             
\end{tabular}
\end{table*}

\begin{table*}[!t]
\centering
\setlength{\tabcolsep}{1.7mm}
\caption{Comparison of cloud segmentation performance across different land cover types in the L8\_Biome dataset.}
\label{tab: l8_biome_scene}
\begin{tabular}{cc}
\begin{tabular}{c|l|cccc}
\toprule
\textbf{Scene}      & \textbf{Method} & \textbf{mIoU} & \textbf{mAcc} & \textbf{mDice} & \textbf{aAcc} \\ \midrule
\multirow{8}{*}{Grass/Crops} 
 & KappaMask~\cite{kappamask} & 36.29 & 62.25 & 42.17 & 76.56 \\
 & CDNetv1~\cite{cdnetv1} & 39.99 & 59.01 & 44.22 & 73.04 \\
 & CDNetv2~\cite{cdnetv2} & 55.17 & 71.84 & 59.23 & 86.78 \\
 & HRCloudNet~\cite{hrcloudnet} & 50.47 & 66.21 & 56.19 & 81.99 \\
 & MCDNet~\cite{mcdnet} & 28.92 & 69.85 & 32.92 & 81.61 \\
 & SCNN~\cite{scnn} & 43.81 & 70.48 & 46.86 & 83.13 \\
 & DBNet~\cite{dbnet} & \textbf{59.08} & 73.85 & \textbf{63.31} & \textbf{89.85} \\
 & UNetMobv2~\cite{cloudsen12_high} & 49.46 & \textbf{74.78} & 54.39 & 88.21 \\
 & Cloud-Adapter~(Ours) & 41.60 & 63.10 & 48.12 & 72.71 \\
\midrule\multirow{8}{*}{Urban} 
 & KappaMask~\cite{kappamask} & 48.83 & 73.39 & 55.55 & 86.78 \\
 & CDNetv1~\cite{cdnetv1} & 46.30 & 64.64 & 51.03 & 81.44 \\
 & CDNetv2~\cite{cdnetv2} & 56.27 & 74.55 & 61.74 & 89.13 \\
 & HRCloudNet~\cite{hrcloudnet} & 56.47 & 71.57 & 62.85 & 87.99 \\
 & MCDNet~\cite{mcdnet} & 31.90 & 70.68 & 36.48 & 83.87 \\
 & SCNN~\cite{scnn} & 48.63 & 71.55 & 52.05 & 85.66 \\
 & DBNet~\cite{dbnet} & \textbf{59.68} & 76.73 & \textbf{65.25} & \textbf{92.14} \\
 & UNetMobv2~\cite{cloudsen12_high} & 50.24 & \textbf{79.90} & 55.39 & 91.59 \\
 & Cloud-Adapter~(Ours) & 51.76 & 76.84 & 58.36 & 86.27 \\
\midrule\multirow{8}{*}{Wetlands}
 & KappaMask~\cite{kappamask} & 40.38 & 61.72 & 47.59 & 75.12 \\
 & CDNetv1~\cite{cdnetv1} & 34.50 & 56.45 & 39.69 & 66.57 \\
 & CDNetv2~\cite{cdnetv2} & 40.54 & 59.62 & 46.71 & 74.78 \\
 & HRCloudNet~\cite{hrcloudnet} & 41.48 & 57.89 & 47.93 & 76.38 \\
 & MCDNet~\cite{mcdnet} & 26.23 & 55.86 & 30.75 & 68.51 \\
 & SCNN~\cite{scnn} & 34.38 & 55.90 & 37.76 & 67.42 \\
 & DBNet~\cite{dbnet} & 49.79 & 67.54 & 56.52 & 83.53 \\
 & UNetMobv2~\cite{cloudsen12_high} & 41.54 & 64.58 & 47.82 & 77.59 \\
 & Cloud-Adapter~(Ours) & \textbf{50.16} & \textbf{74.26} & \textbf{57.09} & \textbf{86.08} \\
\midrule\multirow{8}{*}{Snow/Ice}
 & KappaMask~\cite{kappamask} & 24.41 & 43.24 & 29.37 & 56.74 \\
 & CDNetv1~\cite{cdnetv1} & 20.93 & 38.00 & 23.82 & 44.90 \\
 & CDNetv2~\cite{cdnetv2} & 24.99 & 41.18 & 28.51 & 55.37 \\
 & HRCloudNet~\cite{hrcloudnet} & 22.20 & 36.91 & 25.41 & 47.58 \\
 & MCDNet~\cite{mcdnet} & 10.43 & 25.64 & 13.30 & 29.59 \\
 & SCNN~\cite{scnn} & 12.43 & 23.33 & 14.37 & 24.70 \\
 & DBNet~\cite{dbnet} & 29.72 & 47.25 & 34.04 & 67.69 \\
 & UNetMobv2~\cite{cloudsen12_high} & 23.50 & 41.99 & 27.96 & 56.48 \\
 & Cloud-Adapter~(Ours) & \textbf{32.01} & \textbf{56.99} & \textbf{36.25} & \textbf{73.15} \\
\bottomrule
\end{tabular}
&
\begin{tabular}{c|l|cccc}
\toprule
\textbf{Scene}      & \textbf{Method} & \textbf{mIoU} & \textbf{mAcc} & \textbf{mDice} & \textbf{aAcc} \\ \midrule
\multirow{8}{*}{Barren}
 & KappaMask~\cite{kappamask} & 38.36 & 60.64 & 45.31 & 76.96 \\
 & CDNetv1~\cite{cdnetv1} & 31.50 & 48.05 & 36.15 & 59.96 \\
 & CDNetv2~\cite{cdnetv2} & 41.24 & 59.24 & 46.03 & 76.33 \\
 & HRCloudNet~\cite{hrcloudnet} & 44.33 & 60.53 & 50.19 & 77.51 \\
 & MCDNet~\cite{mcdnet} & 26.56 & 60.52 & 31.34 & 73.00 \\
 & SCNN~\cite{scnn} & 36.93 & 60.42 & 40.55 & 73.35 \\
 & DBNet~\cite{dbnet} & 47.41 & 65.13 & 53.05 & 80.38 \\
 & UNetMobv2~\cite{cloudsen12_high} & 43.66 & 67.80 & 49.67 & \textbf{84.01} \\
 & Cloud-Adapter~(Ours) & \textbf{48.95} & \textbf{72.35} & \textbf{56.45} & 83.49 \\
\midrule\multirow{8}{*}{Forest}
 & KappaMask~\cite{kappamask} & 38.32 & 57.00 & 44.46 & 80.07 \\
 & CDNetv1~\cite{cdnetv1} & 33.86 & 48.50 & 38.67 & 71.81 \\
 & CDNetv2~\cite{cdnetv2} & 41.94 & 58.62 & 47.87 & 81.34 \\
 & HRCloudNet~\cite{hrcloudnet} & 42.93 & 58.30 & 49.64 & 81.76 \\
 & MCDNet~\cite{mcdnet} & 30.05 & 57.39 & 35.23 & 77.71 \\
 & SCNN~\cite{scnn} & 37.62 & 55.44 & 41.39 & 77.92 \\
 & DBNet~\cite{dbnet} & 44.93 & 62.40 & 50.81 & 85.22 \\
 & UNetMobv2~\cite{cloudsen12_high} & 45.23 & 64.89 & 51.17 & \textbf{86.89} \\
 & Cloud-Adapter~(Ours) & \textbf{48.09} & \textbf{68.76} & \textbf{56.25} & 82.30 \\
\midrule\multirow{8}{*}{Shrubland}
 & KappaMask~\cite{kappamask} & 40.87 & 59.71 & 48.28 & 77.26 \\
 & CDNetv1~\cite{cdnetv1} & 32.70 & 49.79 & 38.89 & 61.35 \\
 & CDNetv2~\cite{cdnetv2} & 42.04 & 59.48 & 48.51 & 76.94 \\
 & HRCloudNet~\cite{hrcloudnet} & 41.68 & 57.22 & 48.98 & 76.79 \\
 & MCDNet~\cite{mcdnet} & 28.66 & 57.34 & 34.20 & 71.57 \\
 & SCNN~\cite{scnn} & 37.11 & 56.93 & 41.61 & 71.25 \\
 & DBNet~\cite{dbnet} & 46.49 & 63.41 & 53.76 & 80.88 \\
 & UNetMobv2~\cite{cloudsen12_high} & 41.58 & 63.24 & 48.19 & 79.79 \\
 & Cloud-Adapter~(Ours) & \textbf{51.18} & \textbf{72.01} & \textbf{59.19} & \textbf{87.42} \\
\midrule\multirow{8}{*}{Water}
 & KappaMask~\cite{kappamask} & 42.42 & 64.40 & 49.36 & 81.19 \\
 & CDNetv1~\cite{cdnetv1} & 44.43 & 61.84 & 49.20 & 82.83 \\
 & CDNetv2~\cite{cdnetv2} & 48.72 & 64.17 & 54.10 & 85.07 \\
 & HRCloudNet~\cite{hrcloudnet} & 46.74 & 60.46 & 52.96 & 82.62 \\
 & MCDNet~\cite{mcdnet} & 29.43 & 60.95 & 33.81 & 79.71 \\
 & SCNN~\cite{scnn} & 41.82 & 61.56 & 45.51 & 80.86 \\
 & DBNet~\cite{dbnet} & \textbf{52.42} & 67.85 & \textbf{58.29} & 87.15 \\
 & UNetMobv2~\cite{cloudsen12_high} & 49.58 & \textbf{70.25} & 55.37 & \textbf{88.24} \\
 & Cloud-Adapter~(Ours) & 46.43 & 67.75 & 53.26 & 82.05 \\
\bottomrule
\end{tabular}
\end{tabular}
\end{table*}

\begin{figure*}[!t]
\centering
\includegraphics[width=0.95\linewidth]{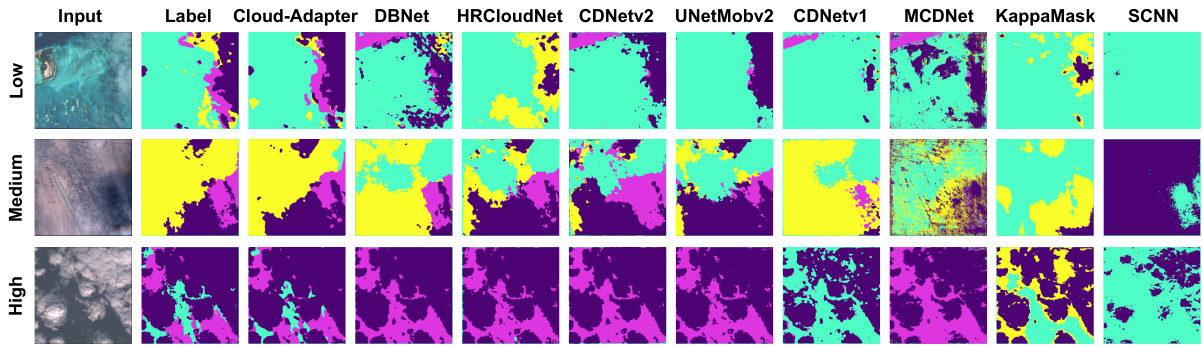}
\begin{tikzpicture}
    \node {
        \begin{tabular}{cccccccc}
            \tikz\fill[clear sky] (0,0) rectangle (0.3,0.3); & Clear Sky &
            \tikz\fill[thick cloud] (0,0) rectangle (0.3,0.3); & Thick Cloud &
            \tikz\fill[thin cloud] (0,0) rectangle (0.3,0.3); & Thin Cloud &
            \tikz\fill[cloud shadow] (0,0) rectangle (0.3,0.3); & Cloud Shadow \\
        \end{tabular}
    };
\end{tikzpicture}
\vspace{-2ex}
\caption{Comparison of visualized segmentation results of different models on the CloudSEN12\_High\_L1C dataset.}
\label{fig: l1c}
\end{figure*}

\begin{figure*}[!t]
\centering
\includegraphics[width=0.95\linewidth]{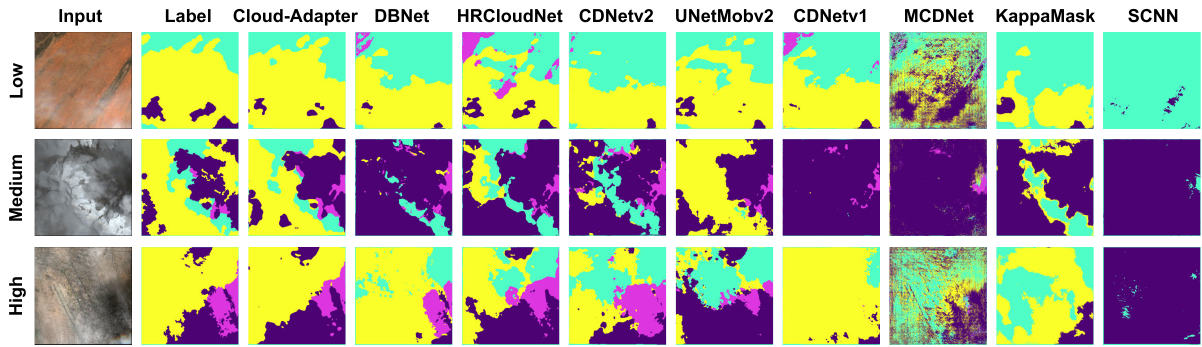}
\begin{tikzpicture}
    \node {
        \begin{tabular}{cccccccc}
            \tikz\fill[clear sky] (0,0) rectangle (0.3,0.3); & Clear Sky &
            \tikz\fill[thick cloud] (0,0) rectangle (0.3,0.3); & Thick Cloud &
            \tikz\fill[thin cloud] (0,0) rectangle (0.3,0.3); & Thin Cloud &
            \tikz\fill[cloud shadow] (0,0) rectangle (0.3,0.3); & Cloud Shadow \\
        \end{tabular}
    };
\end{tikzpicture}
\vspace{-2ex}
\caption{Comparison of visualized segmentation results of different models on the CloudSEN12\_High\_L2A dataset.}
\label{fig: l2a}
\end{figure*}

\begin{figure*}[!t]
\centering
\includegraphics[width=0.95\linewidth]{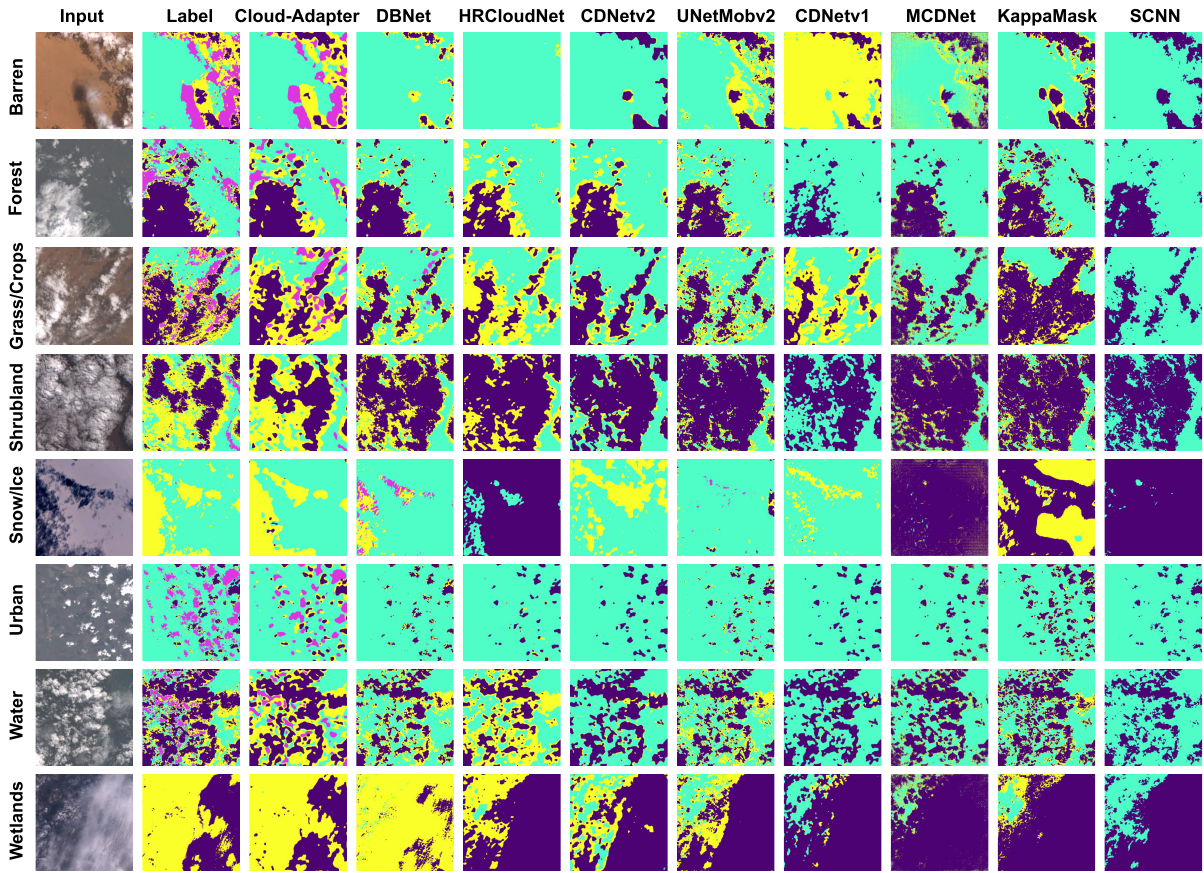}
\begin{tikzpicture}
    \node {
        \begin{tabular}{cccccccc}
            \tikz\fill[clear sky] (0,0) rectangle (0.3,0.3); & Clear Sky &
            \tikz\fill[thick cloud] (0,0) rectangle (0.3,0.3); & Thick Cloud &
            \tikz\fill[thin cloud] (0,0) rectangle (0.3,0.3); & Thin Cloud &
            \tikz\fill[cloud shadow] (0,0) rectangle (0.3,0.3); & Cloud Shadow \\
        \end{tabular}
    };
\end{tikzpicture}
\vspace{-2ex}
\caption{Comparison of visualized segmentation results of different models on the L8\_Biome dataset.}
\label{fig: l8}
\end{figure*}

\paragraph{Quantitative Comparison}

Cloud-Adapter achieves state-of-the-art performance on multi-class cloud segmentation datasets, which include four categories: Clear Sky (CRS), Thin Clouds (TNC), Thick Clouds (TKC), and Cloud Shadows (CDS). On the CloudSEN12\_High\_L1C dataset (Table~\ref{tab: cloudsen12_high_l1c}), Cloud-Adapter achieves a mIoU of 74.18\%, with IoUs of 89.19\% (CRS), 85.46\% (TKC), 56.15\% (TNC), and 65.93\% (CDS). It outperforms methods like UNetMobv2 and HRCloudNet, which show lower scores, particularly in challenging classes such as TNC and CDS. On the CloudSEN12\_High\_L2A dataset (Table~\ref{tab: cloudsen12_high_l2a}), Cloud-Adapter achieves a mIoU of 73.38\%, with IoUs of 89.04\% (CRS), 84.91\% (TKC), 54.97\% (TNC), and 64.60\% (CDS). Its mDice score of 83.87\% further highlights its balanced performance across all categories. On the L8\_Biome dataset (Tables~\ref{tab: l8_biome} and~\ref{tab: l8_biome_scene}), Cloud-Adapter attains the highest mIoU (57.53\%), mAcc (69.34\%), mDice (70.46\%), and aAcc (81.79\%). It excels in challenging categories like Snow/Ice (32.01\%) and Shrubland (51.18\%), consistently outperforming competitors such as DBNet and UNetMobv2 across diverse land scenes.

\paragraph{Visualization Comparison}

Fig.~\ref{fig: l1c} and~\ref{fig: l2a} present Cloud-Adapter's segmentation results on multi-class datasets (CloudSEN12\_High\_L1C and CloudSEN12\_High\_L2A), accurately distinguishing thick clouds, thin clouds, and cloud shadows in complex conditions. The figures also demonstrate its performance across varying cloud cover levels: ``Low" (less than 35\%), ``Medium" (35--65\%), and ``High" (over 65\%). These results highlight Cloud-Adapter's robustness and adaptability to diverse atmospheric scenarios. Fig.~\ref{fig: l8} highlights Cloud-Adapter's segmentation performance on the L8\_Biome dataset, demonstrating superior generalization across diverse biomes, such as barren, forest, and snow/ice regions. These results further emphasize the model’s capability to adapt to varying environmental conditions while maintaining high accuracy and precision and solidify its effectiveness for complex cloud segmentation tasks in remote sensing applications.

In conclusion, Cloud-Adapter consistently outperforms state-of-the-art methods across binary and multi-class datasets, demonstrating its ability to effectively handle diverse cloud segmentation tasks. The visual comparisons further validate its robustness and adaptability, making it a reliable algorithm.

\section{Conclusion}

\noindent{In this paper, we introduced Cloud-Adapter, a parameter-efficient fine-tuning method for cloud segmentation in remote sensing images. By leveraging a frozen VFM and adding a lightweight adapter, Cloud-Adapter reduces the number of trainable parameters while achieving SOTA performance across various datasets. Our approach offers an effective solution for robust cloud segmentation without requiring extensive retraining, making it simultaneously efficient and adaptable.}

\noindent{\textbf{Limitations}. The performance of Cloud-Adapter depends heavily on the VFM, and the large parameters of VFMs make it difficult to deploy on edge devices with limited resources. Additionally, in the complex dataset (L8\_Biome), which encompasses diverse land cover scenarios, our method achieved a leading mIoU score of 57.53\%. However, a substantial gap remains when compared to manually annotated masks. This highlights the potential for improving algorithm efficiency and developing new techniques to enhance segmentation accuracy.}

{
\bibliographystyle{IEEEtran}
\bibliography{refs}
}

\begin{IEEEbiography}[{\includegraphics[width=1in,height=1.25in, clip,keepaspectratio]{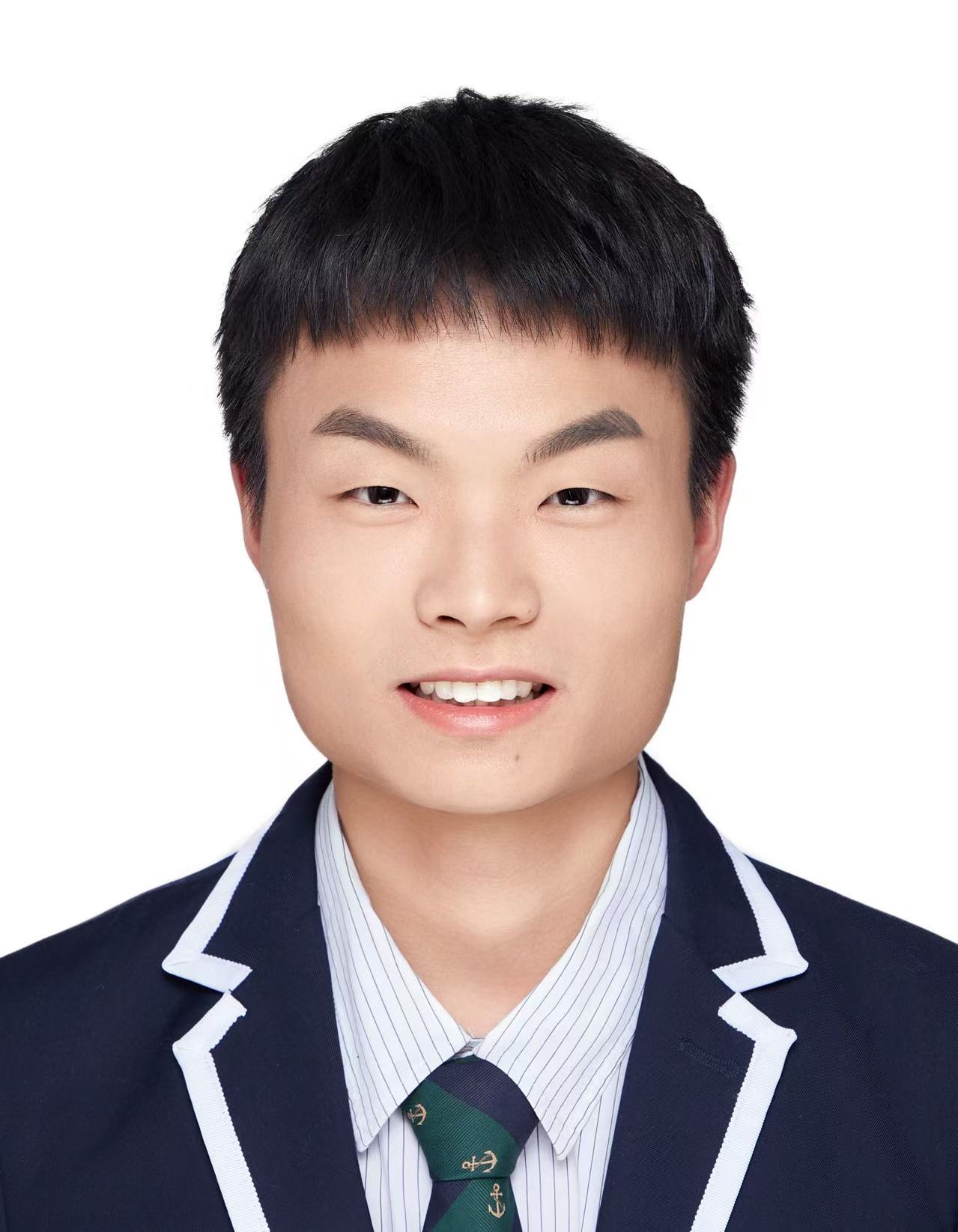}}]{Xuechao Zou} received the B.E. degree in 2021 and the M.S. degree in 2024 from the School of Computer Technology and Application at Qinghai University, Xining, China. He is currently pursuing a Ph.D. degree with the School of Computer Science and Technology at Beijing Jiaotong University. His research interests focus on computer vision, particularly in remote sensing image processing.
\end{IEEEbiography}
\vspace{-8mm}

\begin{IEEEbiography}[{\includegraphics[width=1in,height=1.25in, clip,keepaspectratio]{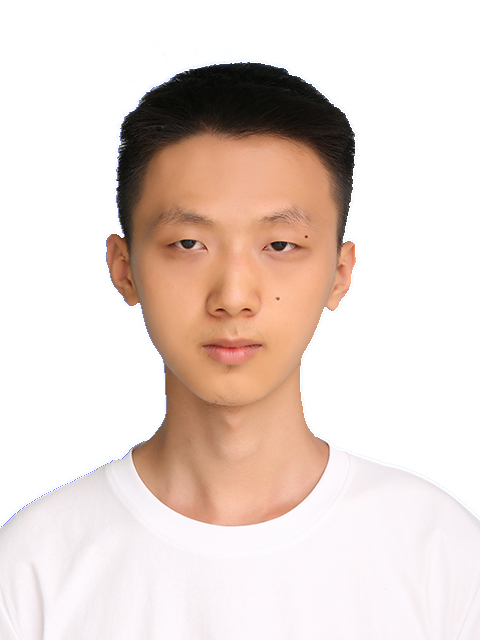}}]{Shun Zhang} is pursuing the B.E. degree with the School of Computer Technology and Application at Qinghai University, Xining, China. His research interests primarily focus on computer vision, particularly in remote sensing image processing.
\end{IEEEbiography}
\vspace{-8mm}

\begin{IEEEbiography}[{\includegraphics[width=1in,height=1.25in, clip,keepaspectratio]{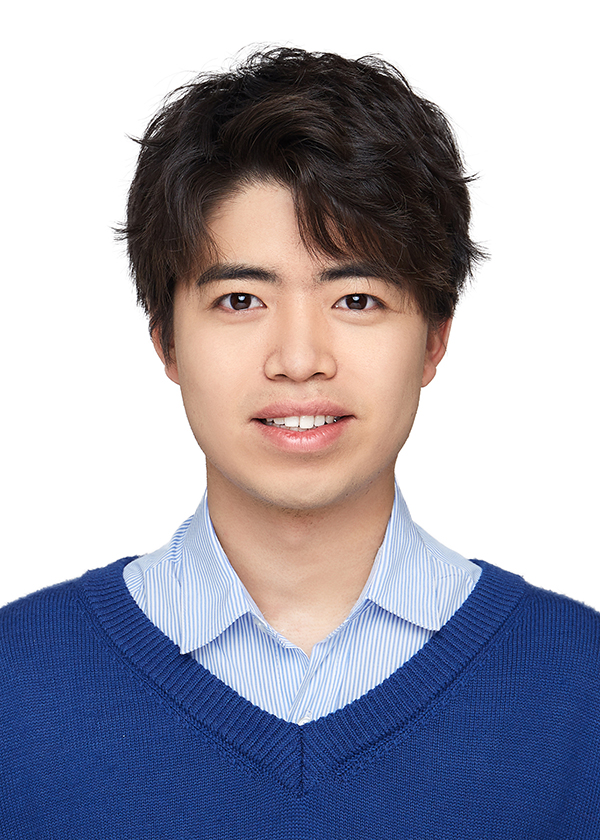}}]{Kai Li} received his B.E. degree from the School of Computer Technology and Application at Qinghai University, Xining, China, in 2020, and his M.S. degree from the Department of Computer Science and Technology at Tsinghua University, Beijing, China, in 2024. He is pursuing a Ph.D. in the Department of Computer Science and Technology at Tsinghua University. His research interests focus primarily on speech separation and audio-visual learning.
\end{IEEEbiography}
\vspace{-8mm}

\begin{IEEEbiography}[{\includegraphics[width=1in,height=1.25in, clip,keepaspectratio]{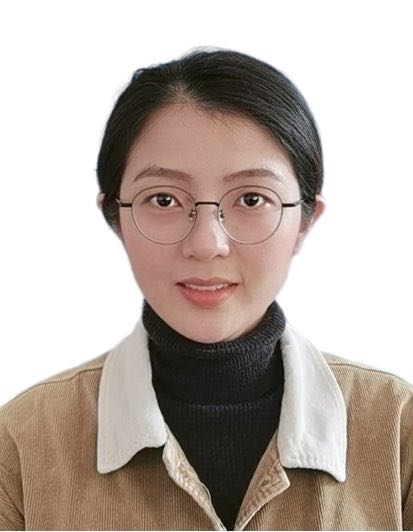}}]{Shiying Wang} received the B.E. degree from Jilin Agricultural University, Changchun, China, in 2015, and the M.S. degree from the School of Computer Technology and Application, Qinghai University, Xining, China, in 2018. She is currently pursuing a Ph.D. degree at Qinghai University. Her research interests mainly include remote sensing image processing, particularly in grassland informatics.
\end{IEEEbiography}
\vspace{-8mm}

\begin{IEEEbiography}[{\includegraphics[width=1in,height=1.25in, clip,keepaspectratio]{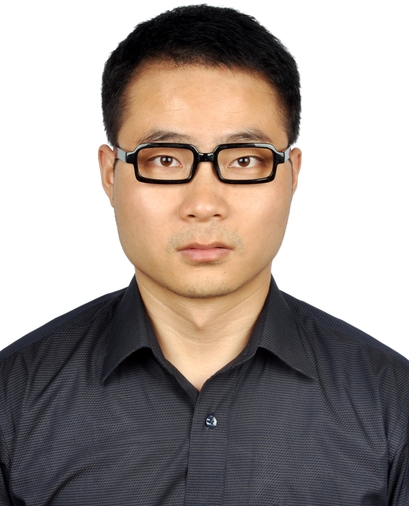}}]{Junliang Xing} received his dual B.S. degrees in computer science and mathematics from Xi’an Jiaotong University, Shaanxi, China, in 2007, and the Ph.D. degree in Computer Science and Technology from Tsinghua University, Beijing, China, in 2012. He is currently a Professor at the Department of Computer Science and Technology at Tsinghua University. He has published over 150 peer-reviewed papers with more than 17,000 citations from Google Scholar. His current research interest focuses on computer vision and gaming problems related to single/multiple agent learning and human-computer interactive learning.
\end{IEEEbiography}
\vspace{-8mm}

\begin{IEEEbiography}[{\includegraphics[width=1in,height=1.25in, clip,keepaspectratio]{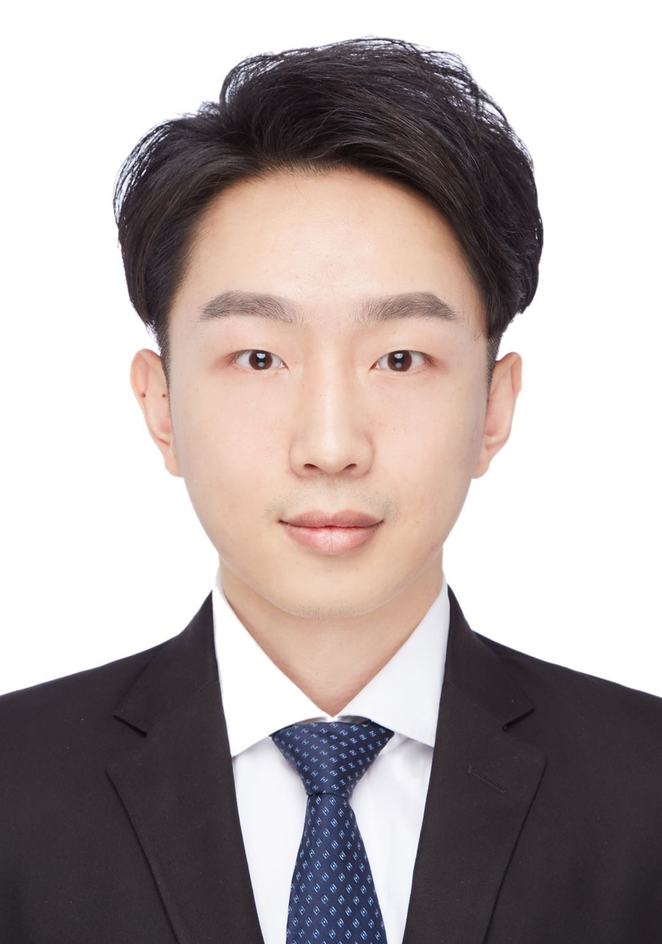}}]{Lei Jin} is currently an associate research fellow at the Beijing University of Posts and Telecommunications (BUPT), Beijing, China. Before that, he graduated from the same university with a Ph.D. degree in 2020. He received a bachelor's degree in the BUPT in 2015. His research interests include machine learning and pattern recognition, focusing on 6D and human pose estimation.  Moreover, he concentrated on network security and analyzed traffic security while pursuing his Ph.D. degree.
\end{IEEEbiography}
\vspace{-8mm}

\begin{IEEEbiography}[{\includegraphics[width=1in,height=1.25in, clip,keepaspectratio]{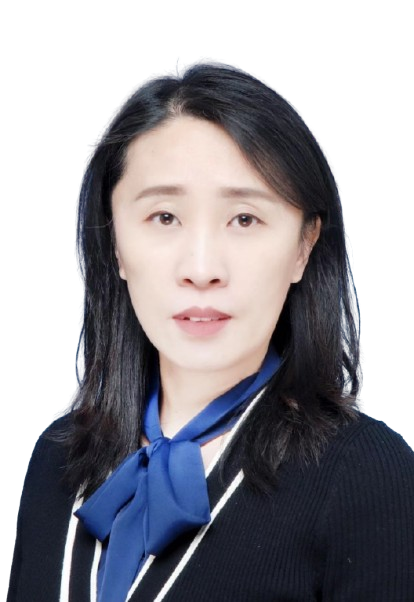}}]{Congyan Lang} received the Ph.D. degree from the Beijing Key Laboratory of Traffic Data Analysis and Mining, School of Computer and Information Technology, Beijing Jiaotong University, Beijing, China, in 2006. She was a Visiting Professor with the Department of Electrical and Computer Engineering, National University of Singapore, Singapore, from 2010 to 2011. From 2014 to 2015, she was a visiting professor at the Department of Computer Science, University of Rochester, Rochester, NY, USA. She is a Professor at the School of Computer and Information Technology, Beijing Jiaotong University. She has published over 80 research articles in various journals and refereed conferences. Her research areas include computer vision and machine learning.
\end{IEEEbiography}
\vspace{-8mm}

\begin{IEEEbiography}[{\includegraphics[width=1in,height=1.25in, clip,keepaspectratio]{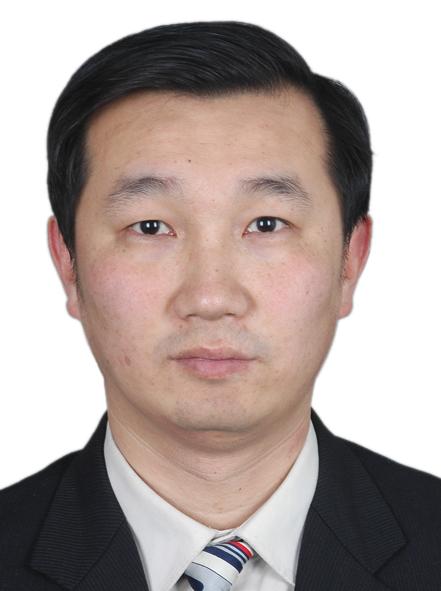}}]{Pin Tao} received his B.S. degree and Ph.D. in computer science and technology from Tsinghua University, Beijing, China, in 1997 and 2002. He is an Associate Professor at the Computer Science and Technology Department, Tsinghua University, Beijing, China. He is also the vice director of the Key Laboratory of Pervasive Computing,  Ministry of Education. Dr. Tao has published more than 80 papers and over 10 patents. His current research interests mainly focus on human-AI hybrid intelligence and multimedia-embedded processing.
\end{IEEEbiography}

\vfill

\end{document}